\pdfoutput=1

\documentclass[11pt]{article}

\usepackage[preprint]{acl}

\usepackage{times}
\usepackage{latexsym}

\usepackage[T1]{fontenc}

\usepackage[utf8]{inputenc}

\usepackage{inconsolata}

\usepackage{soul}
\usepackage{xcolor}
\usepackage{todonotes}
\usepackage{times}
\usepackage{latexsym}
\usepackage[T1]{fontenc}
\usepackage[utf8]{inputenc}
\usepackage{graphicx}
\usepackage{microtype}
\usepackage{enumitem}
\usepackage{inconsolata}
\usepackage{pifont} 
\usepackage{graphicx} %
\usepackage{float} %
\usepackage{afterpage}
\usepackage{placeins}
\usepackage{pifont} 
\usepackage{graphicx} 
\usepackage{array} 
\usepackage{hyperref}
\usepackage{threeparttable}
\usepackage{threeparttablex}
\usepackage{booktabs}
\usepackage{threeparttable}
\usepackage{adjustbox}
\usepackage{pifont}
\usepackage{float}
\usepackage[utf8]{inputenc}
\usepackage{newunicodechar} 
\newunicodechar{≥}{$\geq$}
\usepackage{xcolor}
\usepackage{enumitem}
\usepackage[framemethod=TikZ]{mdframed}
\usepackage{soul, color}
\usepackage{multicol}
\usepackage[most]{tcolorbox}
\usepackage{algorithm}
\usepackage{algpseudocode}
\usepackage{amsmath}
\usepackage{algorithm}
\usepackage{algpseudocode}
\usepackage{multirow} 
\usepackage{amsmath}
\usepackage{algorithm}
\usepackage{algpseudocode}
\usepackage{graphicx}

\newtcolorbox{myprompt}[2][]{
  colback=gray!10,
  colframe=black!60,
  fonttitle=\bfseries,
  colbacktitle=gray!50,
  coltitle=purple,
  enhanced,
  attach boxed title to top left={yshift=-2mm, xshift=5mm},
  title=#2,
  #1,
  sharp corners,
  rounded corners=southeast,
  borderline={0.5mm}{0mm}{dashed},
  boxrule=0.5mm
}

\title{SouLLMate: An Application Enhancing Diverse Mental Health Support with Adaptive LLMs, Prompt Engineering, and RAG Techniques}

\author{Qiming Guo$^{*1}$, Jinwen Tang$^{*2}$, Wenbo Sun$^3$, Haoteng Tang$^4$, Yi Shang$^2$, and Wenlu Wang$^1$ \\
$^1$Texas A\&M University - Corpus Christi, Corpus Christi, TX, USA \\
$^2$University of Missouri, Columbia, MO, USA \\
$^3$Delft University of Technology, Delft, Netherlands \\
$^4$University of Texas Rio Grande Valley, Edinburg, TX, USA\\
$^1$ qguo2@islander.tamucc.edu, wenlu.wang@tamucc.edu\\
$^2$ \{jt4cc, shangy\}@umsystem.edu  $^3$ w.sun-2@tudelft.nl  $^4$ haoteng.tang@utrgv.edu \\}

\begin{document}
\maketitle
\begin{abstract}
Mental health issues significantly impact individuals' daily lives, yet many do not receive the help they need even with available online resources. This study aims to provide diverse, accessible, stigma-free, personalized, and real-time mental health support through cutting-edge AI technologies. It makes the following contributions: (1) Conducting an extensive survey of recent mental health support methods to identify prevalent functionalities and unmet needs. (2) Introducing SouLLMate, an adaptive LLM-driven system that integrates LLM technologies, Chain, Retrieval-Augmented Generation (RAG), prompt engineering, and domain knowledge. This system offers advanced features such as Risk Detection and Proactive Guidance Dialogue, and utilizes RAG for personalized profile uploads and Conversational Information Extraction. (3) Developing novel evaluation approaches for preliminary assessments and risk detection via professionally annotated interview data and real-life suicide tendency data. (4) Proposing the Key Indicator Summarization (KIS), Proactive Questioning Strategy (PQS), and Stacked Multi-Model Reasoning (SMMR) methods to enhance model performance and usability through context-sensitive response adjustments, semantic coherence evaluations, and enhanced accuracy of long-context reasoning in language models. This study contributes to advancing mental health support technologies, potentially improving the accessibility and effectiveness of mental health care globally.

\end{abstract}

\section{Introduction}

The global mental health crisis is a major issue impacting the entire world, affecting an estimated 950 million people and is the leading cause of disability across all age groups. Mental health issues are the leading cause of suicide, with over 800,000 deaths per year\cite{worldwidementalhealth2023}. 
\begin{figure}[h] 
  \centering
  \includegraphics[width=1\linewidth]{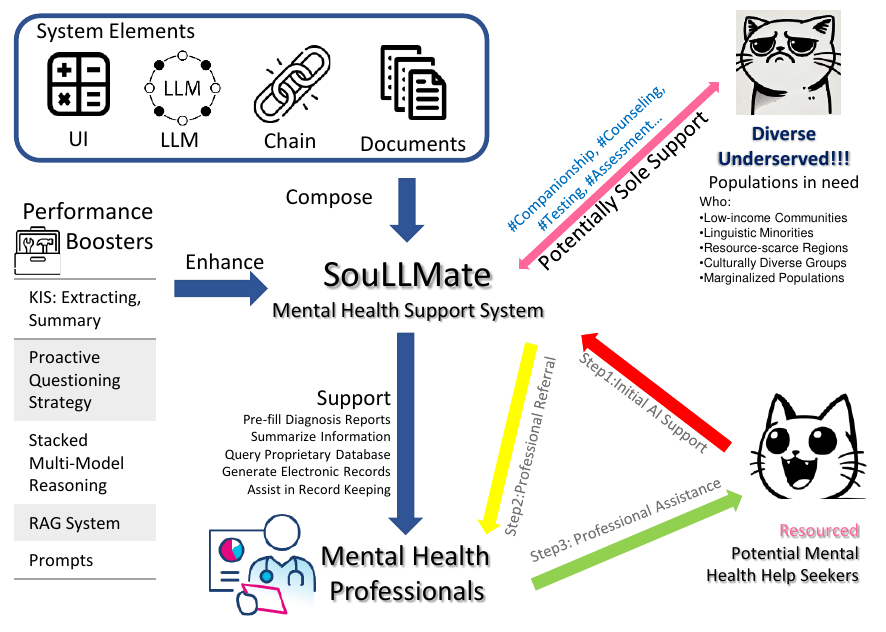}
  \caption{Schematic Diagram of SouLLMate System for Providing Mental Health Support to Diverse Populations}
  \label{fig: 1}
\end{figure}

The crisis is further complicated by a lack of access to mental health services in many parts of the world. According to the National Institute of Mental Health (NIMH), nearly 20\% of adults experience a mental illness each year \cite{NIMH_Mental_Illness}, with the availability of mental health care and support remaining inadequate \cite{bland1997help}. In this context, there is a need for robust mental health support and resources. However, such resources are often concentrated in economically developed urban areas. Even though online mental health support is available, various factors including the cost of care, stigma, and language barriers \cite{bland1997help} hinder access to assistance. Therefore, how to support and assist mental health professionals in making effective diagnoses, as well as how to enhance access to mental health support across different regions, cultural backgrounds, and income levels, remains a crucial issue to address.

\renewcommand{\thefootnote}{}
\footnote{The first two authors * contributed equally to this work.}

To address the issue of limited resources for mental health professionals and the general inadequacy of mental health support available to people, our study introduces SouLLMate, as shown in Figure \ref{fig: 1}, a mental health support system that utilizes LLM technology to relieve mental health professionals from the tedious tasks of collecting patient information, summarizing medical records, and generating reports. This system enables them to dedicate more time and focus more effectively on diagnosis. Additionally, this system can also provide non-clinical support, such as companionship, \cite{galderisi2017proposed, kiesler1979interpersonal} , chat, assessment, and QA functionalities to patients who have already been diagnosed by mental health professionals.

The contributions of our study are:
\begin{itemize}[itemsep=-1pt,parsep=0pt]
    \item We conducted an extensive survey of recent methods supporting mental health, revealing important needs had yet to fulfill.
    
    \item We introduced \textbf{SouLLMate}, as shown in Figure \ref{fig: 1}, an adaptive LLM-driven system for advanced mental health support. This system integrates LLMs, LangChain, RAG, and prompt engineering. SouLLMate offers functionalities beyond General Mental Health Question-Answer and Preliminary Mental Health Assessment, including Risk Detection and Proactive Guidance Dialogue, allow users to upload personal profiles and utilizes Conversational Information Extraction to collect key linguistic indicators, enhancing various support functions, such as Personal Mental Health Reporting.
    
    \item Given the scarcity of evaluation methods for mental health support systems, we developed novel approaches to assess the Preliminary Mental Health Assessment. Our evaluation methodology utilizes professionally annotated interview data and real-life suicide tendency data to rigorously assess these critical functions.
    
    \item We proposed 3 novel methods to enhance model performance and usability: 1) KIS: a context-sensitive adjustment mechanism that modifies responses based on user history and current session data, and 2) PQS: a semantic coherence evaluation strategy ensuring logical conversation flow and sustained user engagement over extended interactions. 3) SMMR: Stacked Multi-Model Reasoning method, designed to enhance the accuracy of long-context reasoning language models.

\end{itemize}

The paper is organized as follows: Background and Investigation; Method; Evaluation; Discussion; Conclusion. Some additional materials are in the Appendix. For further information, see the demo on \href{https://github.com/QM378/SouLLMate}{GitHub} and \href{https://www.youtube.com/watch?v=crxj0VjyFAE}{Video}. Please note that this is a demo system and is continuously being updated.

\section{Background And Investigation}

\subsection{AI for Mental Health}

Artificial intelligence (AI) is increasingly applied across various healthcare domains, including diagnosis and clinical psychology. AI enhances the precision of diagnosing conditions such as COVID-19 through advanced imaging techniques and heart attacks via cardiac monitoring, often matching or surpassing the expertise of clinical professionals \cite{mei2020artificial,han2020ml}.

In the realm of mental health, AI facilitates the diagnosis and treatment of disorders, offers support, generates predictions, and powers supportive chatbots \cite{higgins2023artificial,thieme2020machine,balcombe2023ai,talati2023artificial,tutun2023ai}. Advanced language models, particularly, hold significant promise in bolstering mental health support. For instance, research involving GPT models has assisted healthcare professionals and patients with severe mental conditions in crafting personalized treatment plans \cite{james2023towards,tangshang2024advancing}.

These advancements underscore the robust research and burgeoning application of AI in mental health support, highlighting its increasing importance in healthcare.

\subsection{Key Elements of Effective Mental Health Support Tools}
Recent research underscores the critical need for robust mental health support, particularly for teenagers who frequently turn to online sources before seeking professional help \cite{kruzan2022wanted}. Digital tools with instant messaging and rapid assessments are vital, providing immediate assistance and guiding users to appropriate resources. These features not only promote help-seeking behaviors but also help reduce stigma associated with seeking help—such as time constraints, a preference for self-management, or fear of judgment \cite{czyz2013self}. Effective initial interactions with these tools, marked by timely and relevant support, enhance the likelihood of continued use \cite{frost2016seeks, salaheddin2016identifying}, emphasizing the importance of well-designed digital mental health interventions in sustaining engagement.

\begin{table*}
\vspace{-4mm}
\begin{ThreePartTable}
\label{tab:survey}
\resizebox{\textwidth}{!}{%
\setlength{\tabcolsep}{2pt}
\renewcommand{\arraystretch}{0.85} 
\scriptsize 
\begin{tabular}{l|cccccccccccc}
\hline
Functionalities$\textbackslash$Apps & \textbf{GMH-QA} & \textbf{PMH-A} & \textbf{CIE} & \textbf{SRD} & \textbf{PGD} & \textbf{PMHR} & \textbf{RAG} & \textbf{Lang.} & \textbf{Price} & \textbf{Methods} \\ \hline
Happify \cite{happify2022} & \scalebox{0.75}{\ding{51}} & & \scalebox{0.75}{\ding{51}} & & & \scalebox{0.75}{\ding{51}} & & 10 & \$139.99/Y & IBM technology \\
Headspace \cite{headspace2024} & \scalebox{0.75}{\ding{51}} & & \scalebox{0.75}{\ding{51}} & & & & & 5 & \$69.99/Y & N/A \\
Calm \cite{calm2024} & \scalebox{0.75}{\ding{51}} & & & & & & & 7 & \$69.99/Y & Q bot \\
MindShift \cite{mindshift2024} & \scalebox{0.75}{\ding{51}} & & & & & & & 2 & F+D & GPT 3.5 \\
Shine \cite{shineapp2024} & \scalebox{0.75}{\ding{51}} & & & & & & & 2 & IAP & N/A \\
Bearable \cite{bearable} & \scalebox{0.75}{\ding{51}} & & & & & \scalebox{0.75}{\ding{51}} & & 1 & IAP & N/A \\
Reflectly \cite{reflectly} & & & & & & \scalebox{0.75}{\ding{51}} & & 1 & IAP & N/A \\
Rise Up \cite{riseup} & \scalebox{0.75}{\ding{51}} & & & & & & & 1 & IAP & N/A \\
Rootd \cite{rootd} & \scalebox{0.75}{\ding{51}} & & & & & & & 12 & IAP & N/A \\
Sanvello \cite{sanvello2024} & \scalebox{0.75}{\ding{51}} & & & & & \scalebox{0.75}{\ding{51}} & & 2 & \$199.99/Y & N/A \\
MoodTools \cite{moodtools2024} & \scalebox{0.75}{\ding{51}} & \scalebox{0.75}{\ding{51}} & & & & & & 1 & \$139.99/Y & N/A \\
Breathe2Relax \cite{breathe2relax} & & & \scalebox{0.75}{\ding{51}} & & & & & 1 & Free & N/A \\
Panic Attack \cite{panicattack2024} & \scalebox{0.75}{\ding{51}} & & & & & & & 1 & IAP & N/A \\
Brain.fm \cite{brainfm2024} & & & \scalebox{0.75}{\ding{51}} & & & & & 1 & \$9.99/M & N/A \\
Ada \cite{ada2024} & \scalebox{0.75}{\ding{51}} & & & & & & & 7 & IAP & N/A \\
Amaha \cite{amaha2024} & \scalebox{0.75}{\ding{51}} & & \scalebox{0.75}{\ding{51}} & & & & & 1 & IAP & ML \\
Replika \cite{replika2024} & \scalebox{0.75}{\ding{51}} & & \scalebox{0.75}{\ding{51}} & & & & & 6 & IAP & ML/LLM \\
Serenity \cite{serenity2024} & \scalebox{0.75}{\ding{51}} & & & & & \scalebox{0.75}{\ding{51}} & \scalebox{0.75}{\ding{51}} & 1 & F & GPT 3.5 \\
Woebot \cite{woebot2024} & \scalebox{0.75}{\ding{51}} & & \scalebox{0.75}{\ding{51}} & & & & & 6 & IAP & Gen AI \\
Wysa \cite{wysa2024} & \scalebox{0.75}{\ding{51}} & \scalebox{0.75}{\ding{51}} & \scalebox{0.75}{\ding{51}} & & & & & 1 & IAP & LM \\
Youper \cite{youper2024} & \scalebox{0.75}{\ding{51}} & \scalebox{0.75}{\ding{51}} & \scalebox{0.75}{\ding{51}} & & & \scalebox{0.75}{\ding{51}} & & 2 & IAP & LLM +DL \\
MindDoc \cite{minddoc2024} & \scalebox{0.75}{\ding{51}} & \scalebox{0.75}{\ding{51}} & & & & \scalebox{0.75}{\ding{51}} & & 2 & IAP & N/A \\
Dr. Sila \cite{drsila2024} & \scalebox{0.75}{\ding{51}} & \scalebox{0.75}{\ding{51}} & & & & & & 2 & IAP & N/A \\
Pocketcoach \cite{pocketcoach2024} & \scalebox{0.75}{\ding{51}} & & & & & & & 1 & IAP & N/A \\
\hline
\textbf{SouLLMate}(Our Method) & \scalebox{0.75}{\ding{51}} & \scalebox{0.75}{\ding{51}} & \scalebox{0.75}{\ding{51}} & \scalebox{0.75}{\ding{51}} & \scalebox{0.75}{\ding{51}} & \scalebox{0.75}{\ding{51}} & \scalebox{0.75}{\ding{51}} & 59+ & Local LLM/API usage & LLMs \\
\hline
\end{tabular}%
}

\caption{Functionality comparison between various mental health support systems.}
\label{table:table1}
\begin{tablenotes}
\scriptsize
\item \parbox{0.95\textwidth}{
Functionality Terms: \textbf{GMH-QA} (General Mental Health QA): Common mental health questions and answers.
\textbf{PMH-A} (Preliminary Mental Health Assessment): Initial evaluation of user's mental health.
\textbf{CIE} (Conversational Information Extraction): User information gathering via chat.
\textbf{SRD} (Suicide Risk Detection): Immediate crisis intervention and risk assessment.
\textbf{PGD} (Proactive Guidance Dialogue): Active engagement and guidance based on user input.
\textbf{PMHR} (Personal Mental Health Reporting): Mental health report creation from user data.
\textbf{RAG} (Retrieval-Augmented Generation): Enhanced information retrieval and generation based on user-provided documents and queries.
\textbf{Lang.} (Languages Supported): Number of human languages supported by the system, representing its ability to serve diverse populations.

Price Terms:
\textbf{F} (Free): No cost to use.
\textbf{IAP} (In-App Purchases): Free to download with optional paid features or content.
\textbf{F+D} (Free with Donations): Free to use with optional donations accepted.
*Human-delivered services are excluded for fair comparisons.
}
\end{tablenotes}
\end{ThreePartTable}
\vspace{-4mm}
\end{table*}

\subsection{Investigation of AI Technologies in Mental Health}

To investigate the functionalities that have been supported and the gaps between them and the key elements, we have gathered data through searches of academic papers, web pages, and application markets. The results are presented in Table \ref{table:table1}.

\textbf{Overall, our investigation reveals that, despite some apps performing well in certain aspects, current systems exhibit significant limitations:}

\begin{itemize}[itemsep=-2pt,parsep=0pt]
    \item Unlike human experts who can leverage personal profiles for tailored understanding, most systems lack effective Retrieval-Augmented Generation (RAG) capabilities, diminishing their efficacy and utility.
    \item Human experts initiate proactive dialogue based on individual interests, fostering open communication. In contrast, most existing systems offer only basic, passive question-answering functionalities, often failing to address deep-seated needs.
    \item Human experts can identify and intervene in cases of abnormal tendencies during real-world interactions. However, existing systems largely lack functionalities for detecting such critical conditions.
    \item Current systems and methods lack reliable, evidence-based approaches to validate their effectiveness \cite{nicholas2015mobile, van2015quality}.
\end{itemize}

\section{Method}

\FloatBarrier
\begin{figure*}[h]
  \centering
  \includegraphics[width=1 \linewidth]{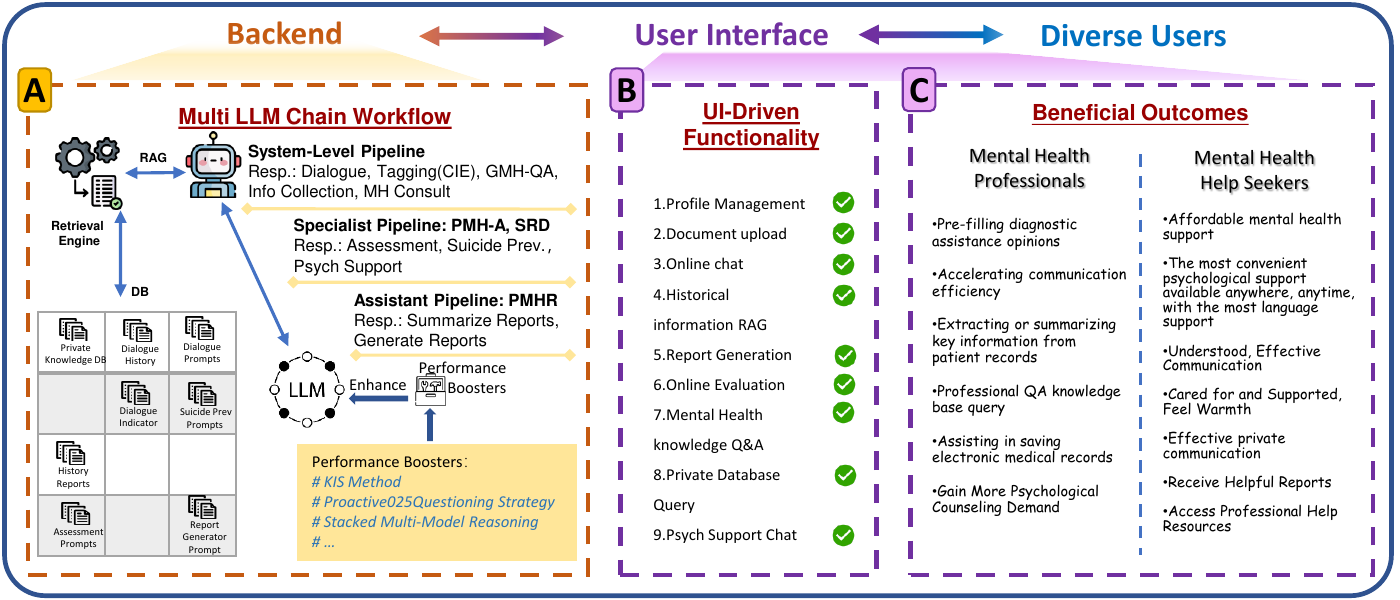}
  \caption{Structure of SouLLMate AI System.}
  \label{fig:workflow}
\end{figure*}

\subsection{System Design}

\textbf{SouLLMate, as shown in figure \ref{fig:workflow}, is designed to assist Mental Health Professionals in diagnosing and providing potential help to Mental Health Help Seekers}, is capable of (1) Preliminary Assessment (PMH-A), (2) obtaining key information through dialogue like humans (CIE), and (3) understanding a person's historical situation or acquiring more document information through reading documents (RAG). (4) Conducting a Proactive Guidance Dialogue (PGD) for individuals who have a negative desire to talk. (5) Risk Detection is also one of the functionalities the system aims to implement. (6) Can generate an appropriate report.

\textbf{To implement these functionalities, we utilize}: Large Language Models (LLMs), LangChain, Retrieval-Augmented Generation (RAG), prompt engineering, and domain expertise. As illustrated in Figure \ref{fig:workflow}, the system is structured into three LLM-based levels:
\vspace{-0.6em} 
\begin{itemize}[itemsep=-2pt,parsep=2pt]
    \item \textbf{System Level}: Implements direct user dialogue, General Mental Health QA (GMH-QA), Conversational Information Extraction (CIE), and RAG functionalities.
    
    \item \textbf{Specialist Level}: Executes Proactive Guidance Dialogue (PGD), Preliminary Mental Health Assessment (PMH-A), and Risk Detection.
    
    \item \textbf{Assistant Level}: Generates the final report.
\end{itemize}
\vspace{-0.5em} 

Due to space constraints, the detailed system design will be presented in the appendix\ref{systemdesign}. 

\subsection{Key Indicator Summarization (KIS)} 

In mental health contexts, which often require long-form dialogues, extended inputs can compromise accuracy and prolong inference times, which cannot be handled by relatively small-size open-source LLM models \cite{sun2024cebench}. To address this, we propose a method that combines prompt engineering with domain expertise to distill key information from historical dialogues. Specifically, we tested and designed two KIS methods: KIS-Summary and KIS-Extracting. KIS-Summary tends to provide a concise yet comprehensive narrative overview, integrating various aspects of health and life into a coherent summary. KIS-Extracting, on the other hand, leans towards precisely extracting and categorizing information in a highly structured manner, facilitating quick retrieval of detailed data for specific categories.

\subsection{Proactive Questioning Strategy (PQS)}

Users often narrate their experiences during conversations, which, while relieving stress and emotions, may not be highly beneficial for the evaluation model. We proposed a proactive questioning strategy, mimicking the inquisitive assessing approach used by psychologists. By pre-defining the LLM's role and behavior in the prompt, the pre-screening process can be expedited, and more relevant information can be gathered. In the SouLLMate system, we designed the PQS (Proactive Questioning Strategy) functionality. Examples of how it works are shown in Table~\ref{tab:example}.
\begin{table}[h!]
  \centering
  \begin{minipage}{\columnwidth}
    \centering
    \setlength{\tabcolsep}{3pt}
    \renewcommand{\arraystretch}{1.2} 
    \small
    \begin{tabular}{m{0.45\columnwidth}|m{0.45\columnwidth}}
      \hline
      \multicolumn{2}{l}{\textbf{User Input:} Today was just one of those days...} \\
      \hline
      Original Response & Response with PQS  \\
      \hline
      Hope tomorrow is better for you! & It sounds like today was tough. Want to talk about what happened? \\
      \hline
    \end{tabular}
    \caption{Response Examples}
    \label{tab:example}
    \vspace{-4mm}
  \end{minipage}
\end{table}

\begin{algorithm}[H]
\caption{Stacked Multi-Model Reasoning (SMMR)}
\label{alg:stacked_multi_model_reasoning}
\begin{algorithmic}[1]
\Require Input data \( X \)
\Ensure Final reasoning result \( R_{\text{final}} \)

\State \textbf{Layer 1: Initial Reasoning}
\State \( R_1 \gets \{M_i(X)\ |\ M_i \in \text{Layer 1}\} \)

\State \textbf{Layer 2 to Layer n: Contextual Reasoning}
\For{\( k = 2 \) to \( n \)}
    \State \( R_k \gets \{M_j(R_{k-1})\ |\ M_j \in \text{Layer } k \} \)
\EndFor

\State \textbf{Layer N: Final Reasoning}
\State \( R_{\text{final}} \gets M_{\text{final}}(R_n) \)

\State \Return \( R_{\text{final}} \)
\end{algorithmic}
\end{algorithm}

\subsection{Stacked Multi-Model Reasoning (SMMR)}

\begin{figure}[h]
    \centering    
    \includegraphics[width=0.4\textwidth]{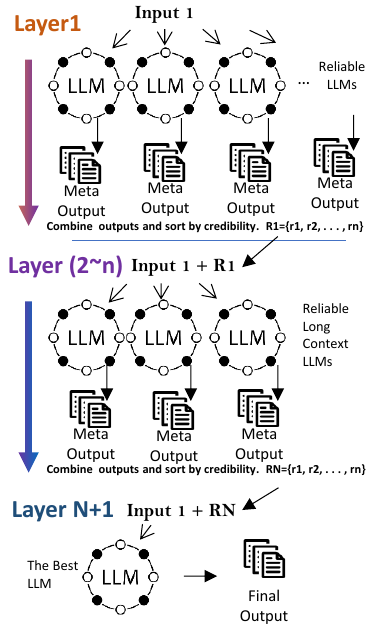} 
    \caption{Structure of the Stacked Multi-Model Reasoning Framework}
    \label{fig:smmr_structure}
\end{figure}

We introduce the concept of Stacked Multi-Model Reasoning (SMMR), specifically designed to enhance the accuracy of long-context reasoning in large language models focused on mental health topics. This approach not only aims to improve reliability but also reduces instability and mitigates the risks of hallucinations often encountered in such models. The architecture of SMMR, structured with a minimum of three layers, is depicted in Figure~\ref{fig:smmr_structure} and Algorithm~\ref{alg:stacked_multi_model_reasoning}.

\vspace{-0.5cm}

\textbf{The Layer 1} of SMMR is composed of multiple reliable single-step reasoning models. Each model operates independently to process the input data \( X \), generating preliminary outputs \(  M_i(X) \). These preliminary outputs are then aggregated as \( R_1 = \{ r_1, r_2, \ldots, r_n \} \).

To ensure the integrity and reliability of Layer 1,o nly those models demonstrating robust single-step reasoning capabilities are selected as Reliable Models for Layer 1, ensuring a strong foundation for subsequent layers.

\textbf{Layer 2 through Layer n}---collectively referred to as the "Middle Layers"---employ models with long-context capabilities designed to synthesize and contextualize the aggregated outputs from the previous layers. Each model $M_j$ within these layers processes the results from the preceding layer (e.g., $R_1$ for Layer 2) to produce refined outputs $r_j$.

The refined outputs from each model in these layers are then aggregated to form the results for that layer, \( R_j \), which serves as the input for the subsequent layer: \( R_j = \{ r_{j1}, r_{j2}, \ldots, r_{jm} \} \).

It is crucial to employ long-context models in Layer 2 through Layer n. This is because tasks in these layers often involve inputs that significantly increase in length, and both smaller Large Language Models and downscaled models tend to exhibit performance declines, especially in long-context situations where they can transform from excellent reasoners to ones prone to hallucinations. For example, while the advanced GPT-4o demonstrates strong single-step reasoning capabilities, its performance in extended contexts is not as robust as that of larger-scale models.

\textbf{The final Layer N} integrates the refined outputs from N-1 Layer using the most advanced model to generate the ultimate reasoning result \( R_{\text{final}} \):

This layer typically employs a single long-context Reliable Model that is most suitable for the current task. It summarizes and consolidates the reasoning results from Layer N-1 to produce the final result.

\section{Evaluation}
\begin{table}[h]
  \centering
  \begin{threeparttable}
    \caption{Mental Health Detection on Case Study Dataset}
    \label{tab:case_study_table}
    \fontsize{8}{11}\selectfont 
    \begin{tabular}{lccccc}
      \hline
      \textbf{Method} & \textbf{Valid (\%)} & \textbf{Acc.} & \textbf{F1} & \textbf{Ave.} & \textbf{SD} \\
      \hline
      GPT-4o       & 92   & 0.95 & 0.98 & 6.97 & 3.20 \\
      GPT-4o+SMMR  & \textbf{100}  & 0.93 & 0.97 & 6.85 & 2.90 \\
      GPT-3.5      & 98   & 0.91 & 0.95 & 6.66 & 3.17 \\
      GPT-3.5+SMMR & \textbf{100}  & \textbf{0.93} & \textbf{0.97} & \textbf{7.03} & \textbf{2.57} \\
      GPT-4        & 100  & 0.92 & 0.96 & 7.02 & 2.95 \\
      GPT-4+SMMR   & 100  & 0.91 & 0.95 & \textbf{7.40} & \textbf{2.67} \\
      \hline
    \end{tabular}
    \begin{tablenotes}
      \small
      \item Note: This table shows the results for both binary mental health detection (accuracy and F1 scores) and the correctness of disorder type identification (Ave, SD).
    \end{tablenotes}
  \end{threeparttable}
\end{table}

\begin{table}[h]
  \centering
  \setlength{\tabcolsep}{1.5pt} 
  {\fontsize{10}{11}\selectfont 
 \begin{tabular}{lccccc}
    \hline
    \textbf{Model} & \textbf{Valid (\%)} & \textbf{F1} & \textbf{Precision} & \textbf{Recall} & \textbf{Acc.(\%)} \\
    \hline
    GPT-4o & 96.57 & 0.86 & 0.83 & 0.90 & 79 \\
    GPT-3.5 & 95.25 & 0.87 & 0.77 & 0.1 & 77 \\
    GPT-4 & 99.21 & 0.89 & 0.81 & 0.98 & 81 \\
    \hline
\end{tabular}
  \caption{Suicide Detection Performance}
  \label{tab:suicide_tendency_accuracy}
  }
\end{table}
\begin{table*}[ht]
  \centering
  \caption{Performance Metrics Across Different Model and Prompts on DAIC-WOZ Dataset}
  \label{tab:prompts}
  \begin{threeparttable}
    \fontsize{9}{10}\selectfont 
    \setlength{\tabcolsep}{1.5pt} 
    \begin{tabular}{lccccccccc}
      \toprule
      \multicolumn{1}{c}{} & \multicolumn{3}{c}{\textbf{GPT-3.5-turbo}} & \multicolumn{3}{c}{\textbf{GPT-4o}} & \multicolumn{3}{c}{\textbf{GPT-4-turbo}} \\
      \cmidrule(lr){2-4} \cmidrule(lr){5-7} \cmidrule(lr){8-10}
      Prompt & Version-1 & Version-2 & Version-3 & Version-1 & Version-2 & Version-3 & Version-1 & Version-2 & Version-3 \\
      \midrule
      MAE & 3.69 & 5.36 & 4.59 & 3.76 & \textbf{2.92} & 6.00 & 4.05 & \textbf{3.15} & 3.71 \\
      RMSE & 5.03 & 6.23 & 5.34 & 5.19 & \textbf{3.94} & 7.95 & 5.58 & \textbf{3.93} & 5.09 \\
      Accuracy & 0.76 & 0.54 & 0.77 & 0.75 & \textbf{0.80} & 0.70 & 0.73 & \textbf{0.78} & 0.78 \\
      Macro-F1 & 0.72 & 0.54 & 0.74 & 0.58 & \textbf{0.78} & 0.41 & 0.54 & \textbf{0.75} & 0.68 \\
      Macro-Precision & 0.71 & 0.65 & 0.73 & \textbf{0.79} & 0.77 & 0.35 & 0.73 & \textbf{0.74} & 0.76 \\
      Macro-Recall & 0.72 & 0.64 & 0.75 & 0.59 & \textbf{0.80} & 0.50 & 0.56 & \textbf{0.77} & 0.66 \\
      ROC-AUC & 0.72 & 0.64 & 0.75 & 0.59 & \textbf{0.80} & 0.50 & 0.56 & \textbf{0.77} & 0.66 \\
      \bottomrule
    \end{tabular}
    \begin{tablenotes}
      \small
      \item Note: For details of the prompts, see the Appendix:~\ref{Appendix:Prompt}.
    \end{tablenotes}
  \end{threeparttable}
\end{table*}
\begin{table*}[htbp] 
\centering
\caption{Performance Enhancement for SMMR Across DAIC-WOZ Subsets}
\label{tab:smmr}
\begin{threeparttable}
\fontsize{10.5}{10}\selectfont 
\setlength{\tabcolsep}{1.5pt} 
\begin{tabular}{l|l|lccccccccc}
\toprule
\textbf{Dataset} & \textbf{Model} & \textbf{Method} & \textbf{Acc.} & \textbf{F1} & \textbf{MF1.} & \textbf{MPrec.} & \textbf{MRec.} & \textbf{ROC AUC} & \textbf{MAE} & \textbf{RMSE} \\
\midrule
\multirow{4}{*}{Training} & \multirow{2}{*}{GPT-3.5-turbo} & P2 & 0.51 & 0.49 & 0.51 & 0.60 & 0.61 & 0.61 & 5.23 & 6.17 \\
                          &                                & P2+SMMR & 0.69 & 0.59 & 0.67 & 0.68 & 0.72 & 0.72 & 4.19 & 5.18 \\
                          & \multirow{2}{*}{GPT-4-turbo}   & P2 & 0.76 & 0.64 & 0.73 & 0.72 & 0.75 & 0.75 & 3.32 & 4.07 \\
                          &                                & P2+SMMR & 0.79 & 0.69 & 0.76 & 0.75 & 0.79 & 0.79 & 3.33 & 4.24 \\
\midrule
\multirow{4}{*}{Testing}  & \multirow{2}{*}{GPT-3.5-turbo} & P2 & 0.55 & 0.57 & 0.55 & 0.70 & 0.68 & 0.68 & 6.04 & 6.81 \\
                          &                                & P2+SMMR & 0.76 & 0.70 & 0.75 & 0.76 & 0.81 & 0.81 & 4.22 & 5.54 \\
                          & \multirow{2}{*}{GPT-4-turbo}   & P2 & 0.77 & 0.67 & 0.74 & 0.74 & 0.77 & 0.77 & 3.21 & 4.16 \\
                          &                                & P2+SMMR & 0.80 & 0.74 & 0.79 & 0.79 & 0.84 & 0.84 & 3.76 & 4.78 \\
\midrule
\multirow{4}{*}{Validation}& \multirow{2}{*}{GPT-3.5-turbo} & P2 & 0.60 & 0.63 & 0.60 & 0.73 & 0.70 & 0.70 & 4.86 & 5.54 \\
                           &                                & P2+SMMR & 0.80 & 0.76 & 0.79 & 0.80 & 0.83 & 0.83 & 3.29 & 3.96 \\
                           & \multirow{2}{*}{GPT-4-turbo}   & P2 & 0.83 & 0.75 & 0.81 & 0.81 & 0.81 & 0.81 & 2.54 & 3.08 \\
                           &                                & P2+SMMR & 0.83 & 0.79 & 0.82 & 0.82 & 0.85 & 0.85 & 2.43 & 3.37 \\
\bottomrule
\end{tabular}
\begin{tablenotes}
\footnotesize
\item Note: P2 refers to Prompt Version 2. Acc, MF1, MPrec, and MRec stand for Accuracy, Macro F1, Macro Precision, and Macro Recall, respectively.
\end{tablenotes}
\end{threeparttable}
\end{table*}
\begin{table*}[htbp]  
\centering
\caption{Comparative Performance Across Models Using Different KIS Strategies on DAIC-WOZ Dataset}
\label{tab:kis_table}
\begin{threeparttable}
\fontsize{9}{10}\selectfont 
\setlength{\tabcolsep}{1pt} 
\begin{tabular}{@{}l l *{7}{c}@{}}
\toprule
\textbf{Metrics} & \textbf{Method} & \textbf{Llama2\_7B} & \textbf{Mistral} & \textbf{Mixtral87B} & \textbf{Llama31\_8b} & \textbf{Llama32\_3B} & \textbf{Llama32\_1B} & \textbf{Gamma2\_9B} \\
\midrule
\multirow{3}{*}{Valid (\%)} 
    & P2           & 2.14  & 28.88 & 34.22 & 28.88 & 25.13 & 10.70 & 8.02 \\
    & P2+KIS1 & 100 & 100 & 100 & 100 & 98.40 & 63.64 & 100 \\
    & P2+KIS2 & 98.93 & 100.00 & 99.47 & 97.86 & 96.79 & 72.73 & 100 \\
\midrule
\multirow{3}{*}{MAE} 
    & P2           & 42.53 & 33.1  & 31.78 & 33.41 & 34.51 & 39.73 & 40.48 \\
& P2+KIS1     & 6.41  & 3.67  & 3.63  & 3.65  & 4.4   & 21.7  & 5.01  \\
& P2+KIS2     & 6.34  & 3.85  & 4.03  & 5.32  & 4.97  & 19.22 & 3.99  \\
\midrule
\multirow{3}{*}{Ave. Runtime (s)} 
    & P2           & 3.04  & 4.04  & 5.91  & 3.04  & 2.73  & 1.63  & 3.55 \\
    & P2+KIS1 & 2.85  & 5.49  & 5.19  & 2.70  & 1.76  & 0.89  & 2.73 \\
    & P2+KIS2 & 2.31  & 4.89  & 4.74  & 2.49  & 1.78  & 0.89  & 2.34 \\
\bottomrule
\end{tabular}
\begin{tablenotes}
\small
\item Note: P2 refers to Prompt Version 2, KIS1 denotes KIS-Summary, and KIS2 denotes KIS-Extracting.
\end{tablenotes}
\end{threeparttable}
\end{table*}
\begin{table*}[ht]
  \centering
  \begin{threeparttable}
    \caption{Model Description and Performance on Domain Knowledge Questions Dataset}
    \label{tab:QA_table}
    \small
    \begin{tabular}{lcccccc}
      \toprule
    \multirow{2}{*}{\textbf{Model Name}} & \multicolumn{2}{c}{\textbf{Short Answer Correctness}} & \multicolumn{2}{c}{\textbf{Multiple Choice Accuracy (\%)}} & \multicolumn{2}{c}{\textbf{Generation Time}} \\
    \cmidrule(lr){2-3} \cmidrule(lr){4-5} \cmidrule(lr){6-7}
     & \textbf{Ave.} & \textbf{SD} & \textbf{488 Questions} & \textbf{100 Questions} & \textbf{Ave.} & \textbf{SD} \\
    \midrule
    GPT-3.5-turbo & 6.48 & 3.13 & 69.88 & 92 & 0.46 & 0.19 \\
    GPT-4-turbo & 6.45 & 3.42 & 72.75 & 94 & 0.55 & 0.35 \\
    GPT-4o & \textbf{7.43} & \textbf{2.77} & \textbf{84.02} & 94 & 0.47 & 0.23 \\
    Mixtral8*7B & 7.11 & 3.04 & 64.34 & 94 & 1.08 & 2.11 \\
    Llama2-7B & 5.85 & 3.09 & 37.50 & 80 & 0.39 & 0.21 \\
    Mistral & 6.98 & 2.99 & 63.52 & \textbf{95} & 0.94 & 1.25 \\
    Gamma2-9B & 6.42 & 3.37 & 69.67 & 94 & 0.34 & 0.21 \\
    Llama3.1-8B & 5.85 & 3.35 & 39.34 & 78 & 0.34 & 0.28 \\
    Llama3.2-3B & 5.76 & 3.09 & 54.30 & 89 & 0.32 & 0.17 \\
    Llama3.2-1B & 4.59 & 2.84 & 29.10 & 54 & 0.28 & 0.12 \\
   \bottomrule
    \end{tabular}
    \begin{tablenotes}
      \small
      \item Note: Performance metrics derived from testing with 100 short answer questions, and two separate multiple choice question sets containing 488 and 100 items respectively.
    \end{tablenotes}
  \end{threeparttable}
\end{table*}
\vspace{-4mm}
\subsection{Evaluation Principle}

As a mental health support system, rigorous evaluation is crucial. A common practice in published AI-driven health support research is to use professionally annotated data as the evaluation sample\cite{aldera2022model,sun2023machine,abunadi2021deep}. 

The meta-analyses conducted by \cite{lecomte2020mobile} found that mental health tools can enhance mental well-being in a promising way. However, few are supported by experimental evidence \cite{neary2018state}. This difficulty stems from the nature of mental health data, which is private and not as easily collectible as other types of data such as images or sounds. The paucity of publicly available datasets for mental health complicates thorough evaluation efforts. To rigorously evaluate the performance of the SouLLMate system, we have devised the following comprehensive evaluation strategy:

1) Collecting high-quality expert-annotated mental health data to assess the system's understanding of mental health issues. If the system demonstrates a high level of understanding and reasoning ability in analyzing cases, it indicates that the system is competent to serve as a general mental health assistant.

2) Gathering domain knowledge questions to test whether the system possesses sufficient and accurate knowledge for answering professional consultation questions \cite{mooney2023towards}.

\subsection{Datasets}

\subsubsection{Professionally Expert-annotated Data} 
\paragraph{The DAIC-WOZ database,}specifically designed for the study of psychological distress such as anxiety, depression, and PTSD, was utilized to assess the performance of our system in mental health pre-screening. This dataset consists of 187 labeled interviews conducted by a virtual interviewer controlled by a human. Each interview is labeled with a PHQ-8 score and its corresponding binary assessment. Training and testing data split IDs are also provided \cite{gratch2014distress}.

In our evaluation, each interview from the dataset was concatenated into a single data stream by aligning segments according to the speaker's starting time. The consolidated dataset includes references to the speaker, the content of each segment, and punctuation marks consisting of a period followed by a slash (./) to denote the end of each speaking turn.\ref{DAIC_data}

\paragraph{The Case Study dataset}compiled by our team from various academic sources including textbooks and journal articles, consists of 48 cases with diagnostic outcomes. For each diagnostic conclusion, binary scores and disorder types were manually extracted to determine the presence of a mental health issue and its specific type. This dataset includes 6 conversational and 42 descriptive cases. An example from this dataset is presented in the Appendix~\ref{Appendix:ExampleOfData}.

\subsubsection{Real World Social Media Data} 
A labeled dataset from Reddit, known as the 'Reddit C-SSRS Suicide Dataset,' was utilized in our study. This dataset, which was created by \citet{gaur2021characterization}, contains 448 entries from Reddit users. These entries are categorized using the Columbia Suicide Severity Rating Scale \cite{posner2011columbia} into Ideation, Behavior, Attempts, Supportive, and Indicator labels. For our analysis, we focused on 379 of these entries, specifically distinguishing between actions associated with suicide risk and supportive actions, while excluding entries labeled as 'Indicator'.

\subsubsection{Domain Knowledge Questions} 
We collected 588 domain knowledge questions, which include about 488 multiple-choice questions and 100 short-answer questions. The multiple-choice questions comprise four columns: ID, question, options, and correct answer. The short-answer questions consist of three columns: ID, question, and answer. These questions were manually selected from professional mental health websites\cite{psychiatric} and books\cite{levy2017case,oltmanns2011case}. Special attention was paid to the publication dates of the materials, and we aimed to select the most recent questions from the second half of 2024 as much as possible.

\subsection{Evaluation of Pre-screening Effectiveness Across Multiple Datasets}
This section details our evaluation of the adaptability and effectiveness of large language models (LLMs) in pre-screening for mental health conditions using three distinct datasets: the DAIC-WOZ database, the Case Study dataset, and the Reddit C-SSRS Suicide Dataset \cite{gaur2021characterization}.

\subsubsection{Prompt Applications}

\paragraph{DAIC-WOZ Database}
We tested three variations of prompts designed to assess the pre-screening capabilities with different models: (Details refer to Appendix:\ref{Appendix:Prompt})
\begin{enumerate}
    \vspace{-0.1cm}
    \item LLMs prompted to evaluate using the Patient Health Questionnaire-8 (PHQ-8) criteria. 
        \vspace{-0.3cm}
    \item LLMs provided with general PHQ-8 criteria for reference.
        \vspace{-0.3cm}
    \item LLMs given detailed item-by-item PHQ-8 evaluation principles. 
        \vspace{-0.5cm}
\end{enumerate} 
We compared PHQ-8 scores from prompts 1, 2, and the cumulative scores from prompt 3, alongside binary classifications (1 for scores ≥ 10). The effectiveness of these prompts is quantitatively summarized in Table \ref{tab:prompts}, where Prompt 2 exhibited superior performance in preliminary screenings.

\paragraph{Case Study Dataset}
We designed prompts that:
\begin{enumerate}
    \vspace{-0.3cm}
    \item Generate a binary score indicating the presence of mental health issues.
        \vspace{-0.3cm}
    \item Identify the type of mental health concern from the content provided.
        \vspace{-0.3cm}
\end{enumerate}
The correctness of the type identifying is auto-evaluated by GPT-4-turbo via a customized 10-point labeled scoring criteria (Detailed in Appendix \ref{Appendix:Prompt}). Results are shown in Table \ref{tab:case_study_table}.

\paragraph{Reddit C-SSRS Suicide Dataset}
Prompts for this dataset aimed to determine:
\begin{enumerate}
    \vspace{-0.3cm}
    \item Whether a post indicates suicidal intentions.
    \vspace{-0.3cm}

    \item If a post suggests behaviors indicative of suicidal attempts.
        \vspace{-0.3cm}
    \item The presence of supportive actions mentioned in the posts.
\end{enumerate}
Responses indicating support are coded as 0; those indicating either intention or behavior, without support, are coded as 1. Results are shown in Table\ref{tab:suicide_tendency_accuracy}.

\subsubsection{SMMR Application}
SMMR was applied to the DAIC-WOZ dataset using the best performing prompt (Prompt 2) to test its effectiveness in enhancing the system's pre-screening performance. The results, demonstrating improved performance cross all subsets, are presented in Table \ref{tab:smmr}. The application of SMMR to the Case Study dataset also showed positive results, detailed in Table\ref{tab:case_study_table}.

\subsubsection{KIS Application}
Two types of KIS strategies, KIS-Extracting and KIS-Summary, were applied to the DAIC-WOZ dataset using the best prompt (Prompt 2) as well. The results, presented in Table \ref{tab:kis_table}, demonstrate that these strategies enable various smaller models to generate more valid and accurate responses in less time.

\subsection{Question-Answer Evaluation and Results}
Domain knowledge questions were assessed using short answer and multiple choice formats in our Domain Knowledge Questions Dataset. The correctness of short answer responses was evaluated based on criteria provided by the LangChain website \cite{LangChain2021}. The proficiency of models like GPT-4o and Mixtral87B was particularly notable, with results highlighted in Table \ref{tab:QA_table}.

\section{Discussion}

\textbf{1 .Benefits to Mental Health Professionals and Patients}

Utilizing SouLLMate or similar systems can significantly alleviate the burden of routine tasks undertaken by mental health professionals. By automating time-consuming activities, these systems allow professionals to dedicate more of their work hours to patient communication and diagnosis. This shift not only enhances the quality of patient interactions but also increases the overall capacity for patient intake without necessitating an increase in the number of mental health professionals. Consequently, this could lead to improved service delivery and potentially higher revenues for mental health practices.

\textbf{2.Adaptability in Multicultural and Multilingual Environments}

Many impoverished and culturally unique regions often lack access to standard psychological counseling services. In such areas, utilizing the internet-based SouLLMate system can provide decent non-clinical support, despite its limitations. This accessibility can facilitate the eventual acquisition of clinical-level mental health support, such as online professional diagnoses. By initially engaging with SouLLMate, individuals may overcome initial hesitations and prejudices towards receiving psychological support, thereby promoting greater acceptance and utilization of mental health services in diverse cultural contexts.

For resource-rich medical institutions, we recommend setting up a server cluster, such as one with eight H100 GPUs, running different models on different GPUs simultaneously. This approach can greatly increase speed when using the SMMR method and enhance the final evaluation and summarization performance.

\textbf{3.What kind of hardware and budget is required to run the system locally, and what are the performance outcomes?}

Currently, the main issue with local models is the inference time; faster inference times mean either stronger hardware or models with lower accuracy, which implies spending more money or reducing performance. In our practice, if a mental health professional wants to establish this system to assist themselves, the cost of local equipment would likely require an A100 server with 80GB of VRAM, which costs approximately USD 10,000 in 2024. Such equipment can run effective models in practice, such as Mixtral8*7B or faster models like llama3 8B. As more advanced models that perform better with the same parameters emerge, the response speed and effectiveness of the SouLLMate system on the same hardware will improve.

For resource-rich medical institutions, we recommend setting up a server cluster that significantly exceeds the performance of the A100, such as with eight H100 GPUs, and running different models on different GPUs simultaneously. This approach can greatly increase speed when using the SMMR method and improve the final evaluation and summarization performance.

\textbf{4.What are the costs associated with using the API?}

Currently, the most effective method is using the API, which charges based on the number of tokens used, meaning you pay for what you use. In our experiment, we estimated that the inference cost for each long conversation in the DAIC dataset is USD 0.06. If each help seeker assisted by a mental health professional requires documentation summarization, report summarization, and report pre-filling functions about 20 times, the cost would be USD 1.20. This work, which SouLLMate completes instantaneously, can save the mental health professional at least 10 minutes.

\section{Conclusion}
Our study developed SouLLMate, an AI-driven system utilizing large language models (LLMs) to support mental health professionals in overcoming limited resources and inadequate support systems. Validated with expert-annotated data, SouLLMate demonstrated strong capabilities achieved an 80\% accuracy rate in clinical mental health assessments under zero-shot conditions. Further improvement can be achieved through methods like SMMR and KIS. This indicates the system’s effectiveness in understanding mental health issues and assisting professionals with tasks such as collecting information, summarizing records, and generating reports. By alleviating these administrative burdens, SouLLMate enables mental health professionals to enhance diagnostic efficiency, serve more patients, and potentially increase their income, creating a win-win scenario for professionals, patients, and service providers.

This research also established baseline performance metrics for this model paradigm, offering a foundation for future studies and providing new insights. We anticipate that integrating more advanced models into systems like SouLLMate will further enhance accuracy and performance. However, SouLLMate faces limitations, such as decreased performance with long contexts, which need to be addressed in future work to enhance its effectiveness and reliability.

\section*{Limitations}

\paragraph{\textbf{Model Constraints:}} Effective mental health pre-screening requires models with high contextual understanding and inference capabilities. However, limitations in model selection restrict the research scope. 

\paragraph{\textbf{Data Scarcity:}} The scarcity of publicly available, professionally annotated data hampers the development and validation of mental health models. The available data is often sparse, unannotated, or excessively lengthy, posing significant challenges. 

\paragraph{\textbf{Resource Costs:}} Mental health research is resource-intensive. Using advanced models like the Mixtral 8*7B on high-end servers can take up to an hour for a single inference, limiting prompt testing and workflow adjustments due to high operational costs.

\section*{Ethics Statement}

\textbf{1. Ethical Boundaries and Professional Support}

SouLLMate’s evaluation principles emphasize rigorous case assessments to determine its expertise and understanding of mental health. The system has achieved favorable evaluation scores in both Q\&A and case study scenarios, indicating its competence as an assistant to mental health professionals. With the advent of more advanced models and improved performance enhancement methods, there remains room for further improvements. SouLLMate is sufficiently professional to serve as an assistant to mental health professionals. \textbf{Importantly, SouLLMate does not engage in scenarios requiring serious diagnoses, thereby maintaining ethical boundaries and ensuring that it complements rather than replaces professional judgment.}

\textbf{2. Privacy and Security of the SouLLMate System}

The SouLLMate system is designed for local deployment, meaning that anyone with access to a capable local large language model or appropriate API services can utilize the system. This design ensures that data and personal information remain under the user’s control or are safeguarded by the large model API service providers. Consequently, SouLLMate itself does not present inherent security issues, as it relies on established security measures provided by the local infrastructure or API providers to protect user data and privacy.

\textbf{3. Privacy and Security of This Research}
Our research does not utilize personal data, relying solely on open-source data and methods, thus avoiding any infringement on personal privacy. When using our system, users have the option to upload or withhold their personal data from the model. Additionally, they can delete the DB file to erase their usage history, thereby protecting their personal privacy.

\section*{Acknowledgements}

This work was partially supported by the National Science Foundation (NSF) under Award No. 2112631, and by the US Department of Transportation (USDOT) Tier-1 University Transportation Center (UTC) Transportation Cybersecurity Center for Advanced Research and Education (CYBER-CARE). (Grant No. 69A3552348332).

\bibliography{custom}

\appendix

\clearpage

\section{Appendix: Deployment and Usage Example}

\subsection{System Deployment}

In this demo, we utilized the simple Panel library to showcase our system. Our demo environment includes Windows 11, Python 3.12, and the EDGE browser, primarily using the Panel library to build the interactive interface that displays the system. Nevertheless, by adjusting the UI and running platform, it has the potential to: \textbf{1)} operate on various platforms and \textbf{2)} demonstrate flexibility that could enable the system to meet diverse needs and consider cultural diversity in its interactions.

\subsection{Use Case for Preliminary Mental Health Assessment}

\label{subsec:Appdendix A.2}
\begin{mdframed}[roundcorner=10pt,backgroundcolor=yellow!5!white!95, linecolor=pink!80!white, middlelinewidth=2pt, innerleftmargin=10pt, innerrightmargin=10pt, innertopmargin=10pt, innerbottommargin=10pt, font=\fontsize{8pt}{10pt}\selectfont]
\textbf{1. Total score:} 20

\textbf{2. Individual scores:}
\begin{enumerate}[itemsep=0pt, parsep=0pt, topsep=0pt, partopsep=0pt]
    \item Lack of interest in activities: 3;
    \item Feelings of depression or hopelessness: 3;
    \item Sleep issues: 3;
    \item Low energy: 3;
    \item Changes in appetite: 1;
    \item Negative self-perception: 2;
    \item Concentration difficulties: 2;
    \item Unusual movement or speech patterns: 3;
\end{enumerate}

\textbf{3. Explanation:}
The individual's diary entry indicates significant challenges with their mental health over the past two weeks. They express a lack of interest in activities they used to enjoy, feelings of depression and hopelessness, sleep issues interfering with their rest, low energy levels, some changes in appetite, negative self-perception, concentration difficulties, and potential signs of anxiety with unusual movement or speech patterns. These scores collectively suggest a high level of distress and potential need for professional support to address these concerns and improve their well-being.
\end{mdframed}
\vspace{-4mm}
\captionof{figure}{One Use Case for Preliminary Mental Health Assessment} 
\label{subsec:figure a1}
\vspace{4mm}
Figure \ref{subsec:figure a1} shows the Preliminary Mental Health Assessment results based on the PHQ-8 protocol. It displays total and item scores with interpretation, demonstrating efficiency and accuracy. This comprehensive output facilitates quick, informed decisions on mental health status.

\FloatBarrier
\begin{figure*}[tp] 
  \centering
  \includegraphics[width=0.95\linewidth]{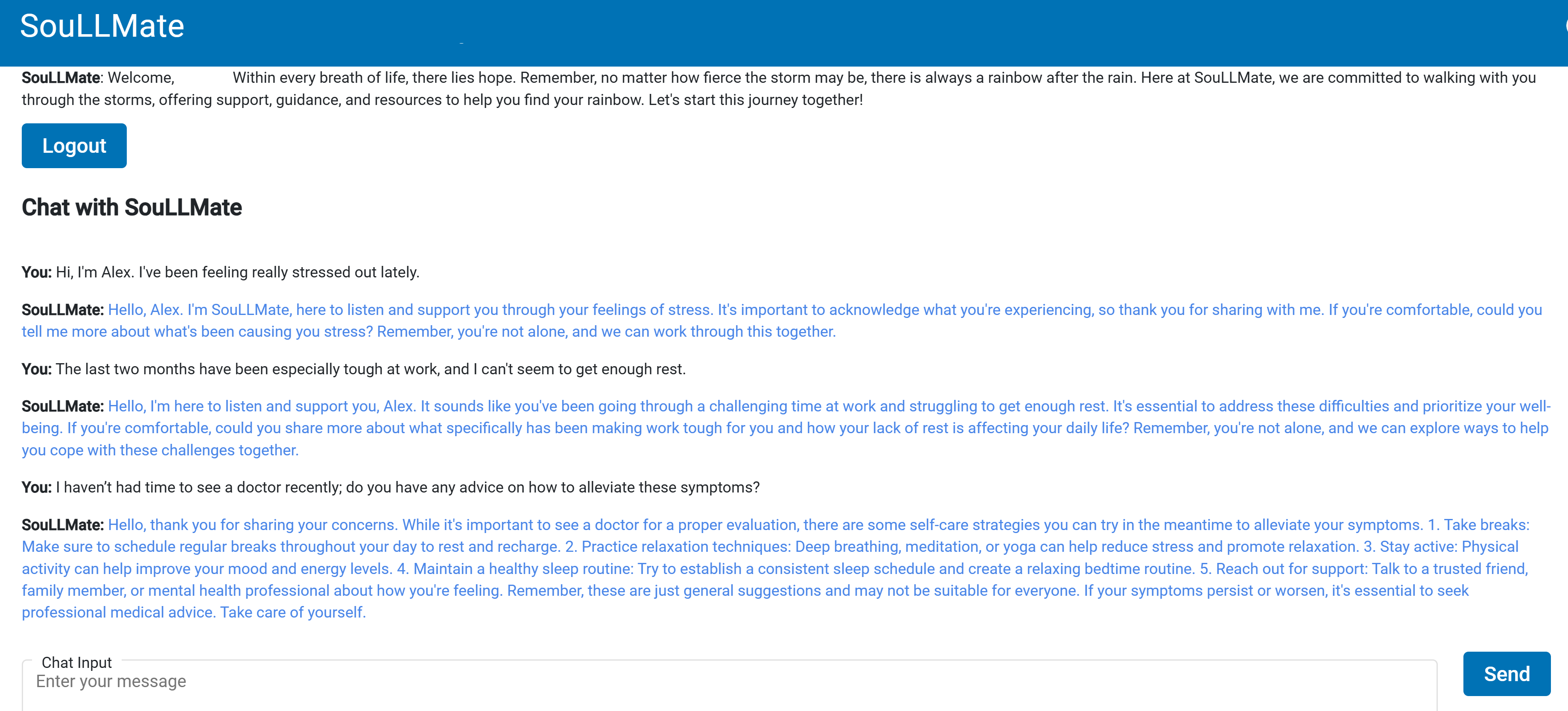}
  \caption{Use Case for General Mental Health QA.}
  \label{fig:Use Case for General Mental Health QA.}
\end{figure*}

\subsection{Use Case for Proactive Guidance Dialogue}

SouLLMate's responses, as shown in Figure \ref{figure:One Use Case for Proactive Guidance Dialogue}, demonstrate empathy, insight, and practical advice. Key aspects include:

\begin{mdframed}[roundcorner=10pt,backgroundcolor=yellow!5!white!95, linecolor=pink!80!white, middlelinewidth=2pt, innerleftmargin=10pt, innerrightmargin=10pt, innertopmargin=10pt, innerbottommargin=10pt, font=\fontsize{8pt}{10pt}\selectfont]

\textcolor{blue}{You:} I guess I'm here because I sometimes feel a bit overwhelmed. It's not all the time, just now and then.

\textcolor{orange}{SouLLMate:} Hello, it's completely normal to feel overwhelmed at times, especially with everything going on in our lives. \textcolor{purple}{It's great that you're aware of these feelings and willing to talk about them. Remember, it's okay to reach out for help when you need it, whether that's talking to someone you trust or seeking professional support. You're not alone in this, and we can work together to find healthy ways to manage those overwhelming moments.}[\textsuperscript{[1]}] Take care of yourself.

\textcolor{blue}{You:} Lately, I find it harder to enjoy things I used to like; not sure why that's happening.

\textcolor{orange}{SouLLMate:} Hello, it's understandable to feel this way when you notice a decrease in your enjoyment of activities you used to like. \textcolor{purple}{Changes in our interests can be influenced by various factors, including stress, mood, or life circumstances. It's great that you're reflecting on these feelings and seeking insight into why it might be happening.}[\textsuperscript{[2]}] Remember, it's okay to explore these emotions further and consider reaching out for support if needed. You're not alone, and we can work together to understand and address these changes. Take care of yourself.

\textcolor{blue}{You:} At night, I often find myself thinking about things I didn't finish during the day. It keeps me up.

\textcolor{orange}{SouLLMate:} Hello, it's common to experience racing thoughts at night, especially when reflecting on unfinished tasks from the day. \textcolor{purple}{This can make it challenging to relax and fall asleep. One helpful strategy is to create a bedtime routine that includes wind-down activities like reading or gentle stretching to signal to your body that it's time to rest. Writing a to-do list before bed can also help clear your mind of lingering thoughts.}[\textsuperscript{[3]}] Remember, it's essential to prioritize your rest for overall well-being. If these thoughts persist, consider speaking to a mental health professional for additional support. Take care of yourself.

\textcolor{blue}{You:} Sometimes, I wonder if it’s normal to feel this way or if I should be doing something about it.

\textcolor{orange}{SouLLMate:} Hello, it's completely normal to have these thoughts and feelings, especially when you're experiencing challenges with your mental health. \textcolor{purple}{It's essential to acknowledge what you're going through and consider seeking help if you feel it's impacting your well-being. You're taking a positive step by reflecting on these concerns, and it shows strength to consider ways to address them.}[\textsuperscript{[4]}] Remember, it's okay to prioritize your mental health and explore support options that can help you navigate these emotions. You're not alone, and we can work together to find ways to improve your well-being. Take care of yourself.

\end{mdframed}
\vspace{-4mm}
\captionof{figure}{One Use Case for Proactive Guidance Dialogue} 
\label{figure:One Use Case for Proactive Guidance Dialogue}

\vspace{-1mm}

\begin{itemize}[itemsep=-2pt,parsep=2pt]
\vspace{-4mm}
\item [\textcolor{purple}{1}] \textbf{Empathetic Validation:} Normalizing emotions and encouraging disclosure.
\item [\textcolor{purple}{2}] \textbf{Cognitive Exploration:} Guiding self-reflection on emotional states.
\item [\textcolor{purple}{3}] \textbf{Behavioral Intervention:} Offering strategies for stress-induced insomnia.
\item [\textcolor{purple}{4}] \textbf{Psychoeducation:} Promoting proactive mental health management.
\end{itemize}
\vspace{-4mm}
These interactions illustrate SouLLMate's ability to provide comprehensive psychological support through evidence-informed dialogues.

\subsection{Use Case for Suicide Risk Detection}

As shown in Figure \ref{figure:Use Case for Suicide Risk Detection}, this assessment effectively evaluates suicide risk by quantifying key factors such as intent, ideation, and behavior. It accurately identifies the absence of suicidal tendencies while recognizing the supportive nature of the response, demonstrating its efficacy in mental health dialogues.

\begin{mdframed}[roundcorner=10pt,backgroundcolor=yellow!5!white!95, linecolor=pink!80!white, middlelinewidth=2pt, innerleftmargin=10pt, innerrightmargin=10pt, innertopmargin=10pt, innerbottommargin=10pt, font=\fontsize{8pt}{10pt}\selectfont]
\textbf{Content:} "It's not a viable option, and you'll be leaving your wife behind. You'd pain her beyond comprehension. It sucks worrying about money, I know that firsthand. It can definitely feel hopeless, as you seem to be tiredly aware of. Your wife might need to chip in financially. I know time is an issue, but even 10-15 hours a Asthenia could alleviate a lot of the pressure. In the meantime, get your shit together - write that resume tomorrow. No excuses, get it done and send it out. Whether you believe in some sort of powerful being or force governing things or not, things really do work themselves out. This is a big test for you, and youll pull through. Just try to stay as positive as you can and everything will work out."

\textbf{Generated Results:}
\begin{itemize}[itemsep=-2pt,parsep=2pt]
\vspace{-3mm}
    \item Suicide intent: 0
    \item Suicide phrase: NA
    \item Passive Ideation: 0
    \item Active Ideation: 0
    \item Intent: 0
    \item Plan: 0
    \item Behavior: 0
    \item Supportive: 1
    \item User: user-0
\end{itemize}
\end{mdframed}
\vspace{-6mm}
\captionof{figure}{Use Case for Suicide Risk Detection} 
\label{figure:Use Case for Suicide Risk Detection}

\subsection{Use Case for Report Generation}

\vspace{-5mm}
\FloatBarrier
\begin{figure}[H] 
  \centering
  \includegraphics[width=0.95\linewidth]{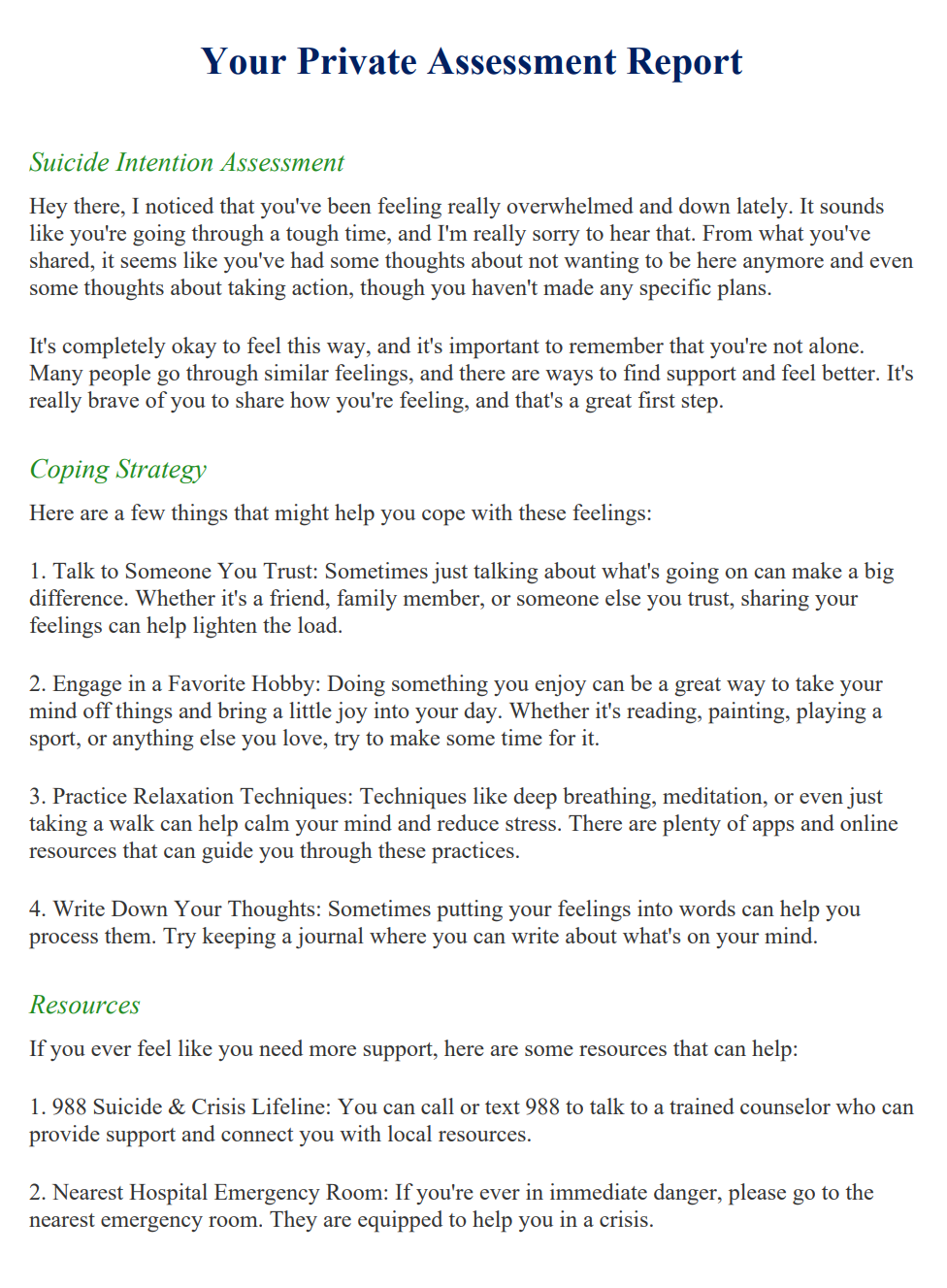}
  \caption{Use Case for Report Generation.}
  \label{fig:Use Case for Report Generation.}
\end{figure}

\subsection{Use Case for General Mental Health QA}
\label{subsec:Appdendix A.3}

The SouLLMate, as illustrated in Figure \ref{fig:Use Case for General Mental Health QA.}, excels in empathy and active listening, offering practical stress relief advice while maintaining professional boundaries. Its responses are well-structured and personalized, using supportive language like "You're not alone." 

\section{Appendix: Detailed Design of SouLLMate}
\label{systemdesign}

\subsection{Design of System Level Workflow}

The System Level workflow is designed as a cyclical dialogue chain. It serves as the initial direct contact enabling the SouLLMate system to connect with the user, after which other components can operate accordingly.

\textbf{\textbullet} Responsibility: The workflow is tasked with engaging in dialogue with users; identifying and extracting indicators and information through conversation; connecting to the database which stores pre-installed knowledge-based documents along with users' historical information and customized documents; and conducting General Mental Health QA (GMH-QA) by invoking RAG to access and utilize database information.

\textbf{\textbullet} Prompt Design: The prompts primarily define the LLM's role as a system-level LLM, outlining its methods and styles of dialogue with users, authority in responses, and the rights to invoke RAG and access or update the user's state.

\textbf{\textbullet} RAG Design: The moment a user interacts with the system, the RAG system is invoked at the System Level to retrieve user history and directly contribute to the dialogue and consultation.

\subsection{Design of Specialist Level Workflow}

The specialist workflow consists of three components, which implement PGD, PMH-A, and SRD functionalities based on the information obtained from the System Level LLM.

\paragraph{Proactive Guidance Dialogue (PGD)} 
\hspace{\parindent} \textbf{\textbullet} Responsibility: PGD proactively guides dialogue, helps users alleviate emotions, and encourages openness. It utilizes RAG functionality to access the database (The database contains knowledge documents, the user’s most recent emotional state, historical assessment data, and past suicidal tendencies).

\textbf{\textbullet} Prompt Design: The prompt primarily defines the LLM's role as an expert, patiently discovering users' mental health needs and actively guiding topics to encourage users to express themselves and open up.

\paragraph{Preliminary Mental Health Assessment (PMH-A)}
\hspace{\parindent} \textbf{\textbullet} Responsibility: PMH-A evaluates user-submitted assessment documents or conducts dialogue-based assessments in the current conversation window. Based on the defined assessment prompts, utilize the Patient Health Questionnaire–8 (PHQ-8) \cite{Kroenke2002} to assess and score the obtained text, identifying scores ≥10 as indicative of mental health issues.

\textbf{\textbullet} Prompt Design: The prompt primarily defines the LLM's role as an assessment expert, the questioning strategy in user dialogues, the PHQ-8 scoring criteria for evaluations, and the format for producing assessment outcomes.

\paragraph{Suicide Risk Detection (SRD)}

\hspace{\parindent} \textbf{\textbullet} Responsibility: This functionality assesses users' suicidal tendencies using historical data, documents in the database, user dialogues, and uploaded information accessed through RAG. It also evaluates these tendencies according to three professional dimensions: intention, behavior, and whether the individual has received support from others. 

\textbf{\textbullet} Prompt Design: The prompt primarily defines the LLM's role, establishes the criteria and basis for utilizing suicidal tendency detection through stages such as intent, ideation, behavior, and plan.

\subsection{Design of Assistant Level Workflow}

\hspace{\parindent} \textbf{\textbullet} Responsibility: Create personalized reports for users to download and keep as records, or they can submit these to mental health professionals to provide pre-screening information and historical records.

\textbf{\textbullet} Prompt Design: The prompt primarily defines the LLM's role as a summarizer and advisor, tasked with generating a comprehensive report based on users' recent assessment results, dialogue history, and current status.

\section{Example of Data}
\label{Appendix:ExampleOfData}

Case Study Dataset Example:
Content: A 37-year-old white male infantryman stationed in Iraq arrived at a field hospital complaining that his superior officer placed poisonous ants in his helmet. His face is covered with excoriations from persistent scratching. On further examination, he is stuporous and has mildly slurred speech, tremor, and mint odor to his breath. Later his troop leader mentioned that his Humvee was littered with empty bottles of mouthwash and that the man has been reprimanded for falling asleep at his post. After a night of rest, he discussed his excessive use of mouthwash in place of alcohol, which is the only available form of alcohol in Iraq.

Conclusion:  The individual in the provided case study exhibits symptoms consistent with delusions of persecution (believing his superior officer placed poisonous ants in his helmet), excoriations from persistent scratching indicative of possible hallucinations or delusions, stupor, slurred speech, tremors, and the smell of mint on his breath. Additionally, his excessive use of mouthwash as a substitute for alcohol suggests a coping mechanism or self-medication behavior. While the text does not explicitly provide a diagnosis, the symptoms described align with features of psychosis and substance abuse, possibly indicating comorbid conditions such as schizophrenia spectrum disorder and alcohol use disorder.

\section{Appendix: KIS Example}
\subsection{KIS Summary}
\label{Appendix:KIS_summary}
**Demographics**: The participant is from Los Angeles, currently living in West Los Angeles.

**Health History**: There is no mention of significant physical health issues. The participant has not been diagnosed with depression or PTSD and has not served in the military.

**Mental Health Status**: The participant expresses a generally stable emotional state, with no high highs or low lows. They mention being generally happy and content with daily achievements. However, they also express frustration with people who provoke them and a desire to handle stress and temper better.

**Coping Mechanisms**: The participant engages in physical activities like running and going to the gym to relax. They also enjoy music and spending time in creative environments, which suggests a healthy approach to stress management.

**Risk Factors**: The participant indicates some frustration with their current job situation and the general pressures of living in a competitive city like Los Angeles. They also show sensitivity to provocation by others, which can be a source of stress.

**Major Health Concerns**: The main concerns are related to stress management and emotional regulation, particularly in response to provocation.

**Treatment History**: There is no mention of any ongoing or past mental health treatments.

**Immediate Help**: There is no indication that immediate help or intervention is necessary based on the conversation.

**Conversation Context**: The conversation appears to be a structured interview or assessment, possibly in a therapeutic or research setting, aimed at understanding the participant's lifestyle and mental state.

**Lifestyle Overview**: The participant is active, enjoys creativity, and values independence in their work. They live alone and seem to have a supportive social network, although they do not mention a best friend specifically. They are close to their family and find inspiration from friends who are doing well.

Overall, the participant appears to be managing daily stresses reasonably well, with a good support system and healthy coping mechanisms, though they could benefit from further developing strategies to manage provocations and stress more effectively.

\subsection{KIS Extracting}
\label{Appendix:KIS_extracting}
Demographics: from Los Angeles, lives in West Los Angeles, studied business, works as an administrative assistant through a temp agency

History: NA

Positive Indicator: "I'm doing good thank you", "I love it I'm from here", "I'm pretty close [to my family]", "I'm usually pretty I feel like I'm a level person and I'm pretty happy"

Negative Indicator: "the traffic is horrible", "I feel like I could do more", "stupid people just basically doing just doing anything to annoy me", "people provoking me"

Coping and Regulation: tries not to remember bad situations, tries not to let provocations get to him, likes to run, go to the gym, listen to music, spends time around musicians

Risk Factors: lives alone, feels he could do more than his current job, experiences provocation from others, traffic stress

Major Concern: dissatisfaction with current job situation, dealing with provocation from others

Treatment: NA

Immediate Help: No

Conversation Type: formal interview

Summary: The participant is a resident of West Los Angeles who has studied business and currently works as an administrative assistant through a temp agency. He enjoys the local weather and the vibrant city life, although he dislikes the traffic. He lives alone and values his close relationship with his family. His main recreational activities include running, going to the gym, and engaging with music and art. He expresses some dissatisfaction with his current job and has faced challenges with people provoking him, which he tries to manage by not dwelling on these incidents. He does not report any major regrets or diagnosed mental health issues. Overall, he maintains a level-headed approach to life's ups and downs, focusing on daily accomplishments to stay positive.

\section{Appendix: Prompts}
\label{Appendix:Prompt}
In this section, we showcase the prompts used in the paper for implementing various functions, along with the criteria prompts for scoring. More prompts, aimed at extracting historical information and supporting RAG and multi-agent processes, can be found in our supplementary code. It must be noted that while the prompts presented in the paper may not seem numerous or complex, the design and finalization of the prompts currently in use involved extensive experimentation and testing. Different models respond differently to the same prompt; some perform better with shorter prompts, while others excel with more detailed ones. Some models completely fail to support long prompts, possibly limited by the available context, showing a trend where longer inputs result in poorer performance. Others are highly sensitive to the input, displaying a tendency to refuse making decisions when faced with even slightly uncertain information. Therefore, although we have displayed the prompts we used in the appendix, they are not universally applicable across different models.


\begin{figure*}[!h] 
  \centering
  \includegraphics[width=0.9\linewidth]{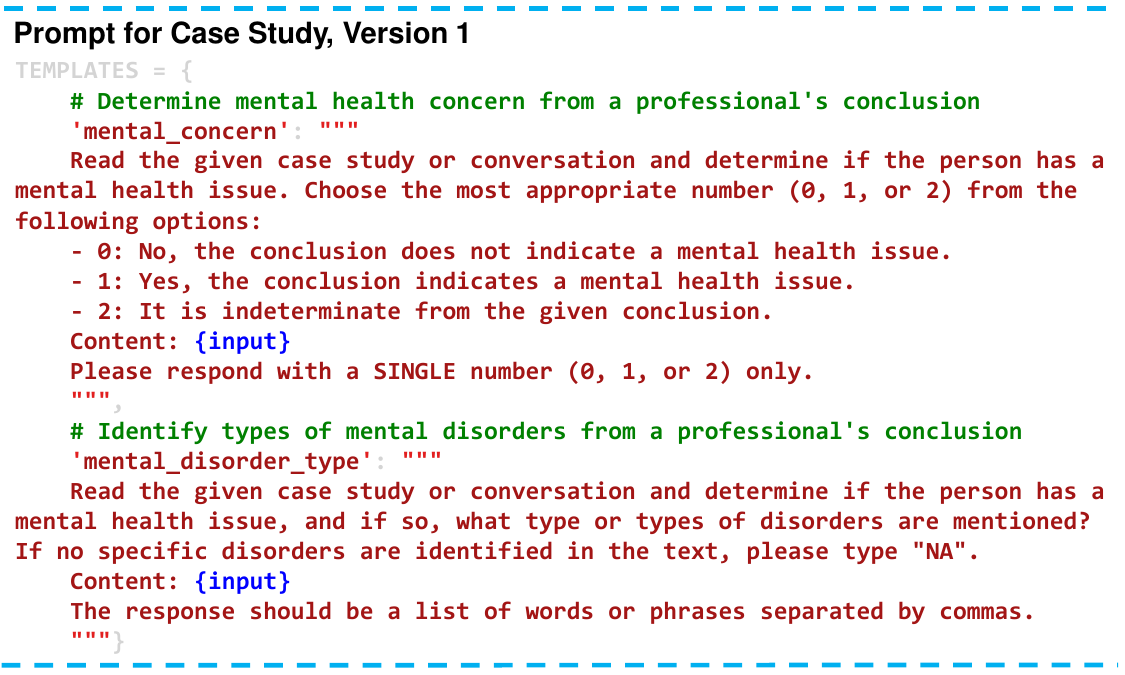}
  \caption{Prompt for Case Study, Version 1}
  \label{fig:Prompt for Case Study, Version 1}
\end{figure*}


\begin{figure*}[!h] 
  \centering
  \includegraphics[width=0.9\linewidth]{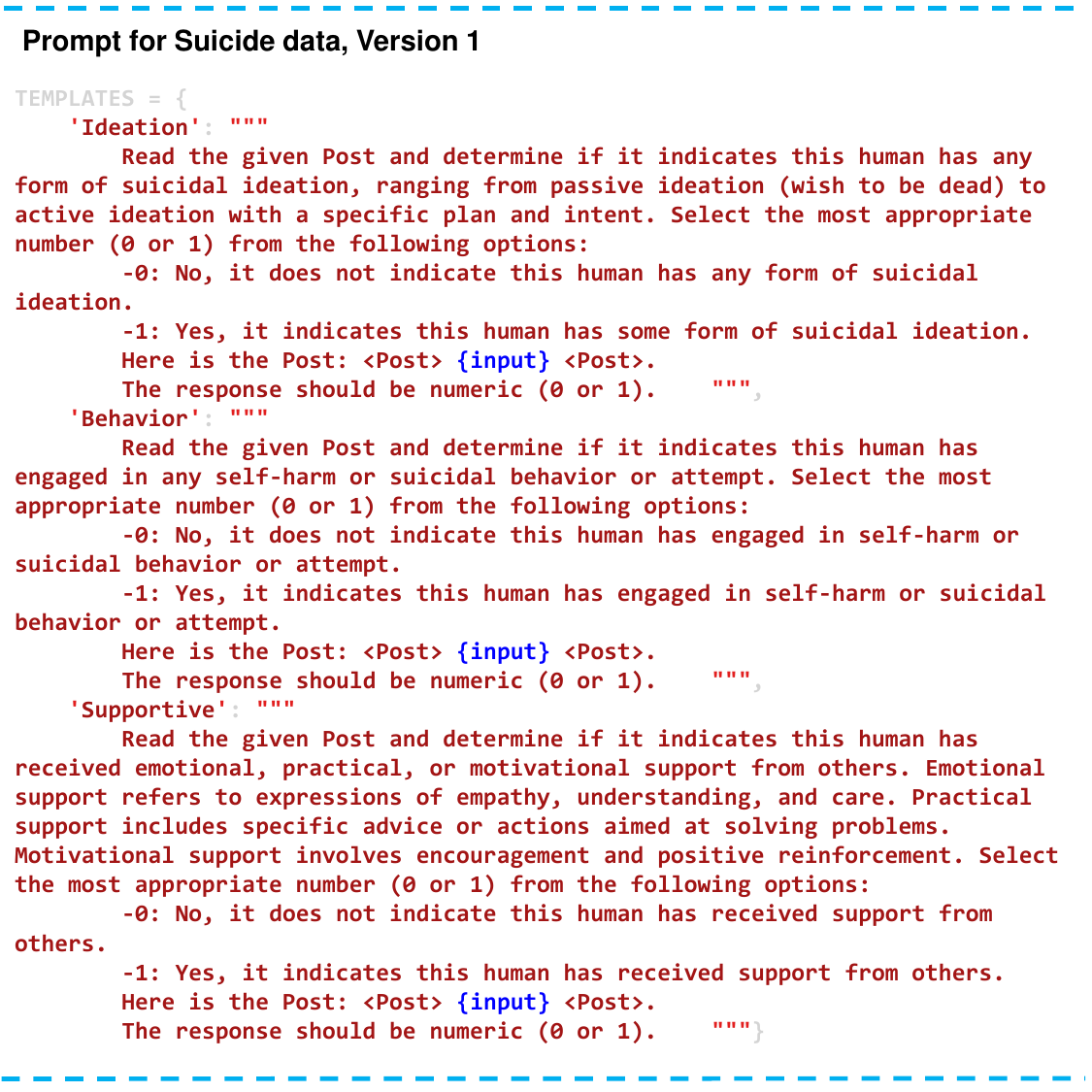}
  \caption{Prompt for Suicide Data Version 1}
\label{fig:Figure_P_Suicide_data, Version 1}
\end{figure*}


\begin{figure*}[!h] 
  \centering
  \includegraphics[width=1\linewidth]{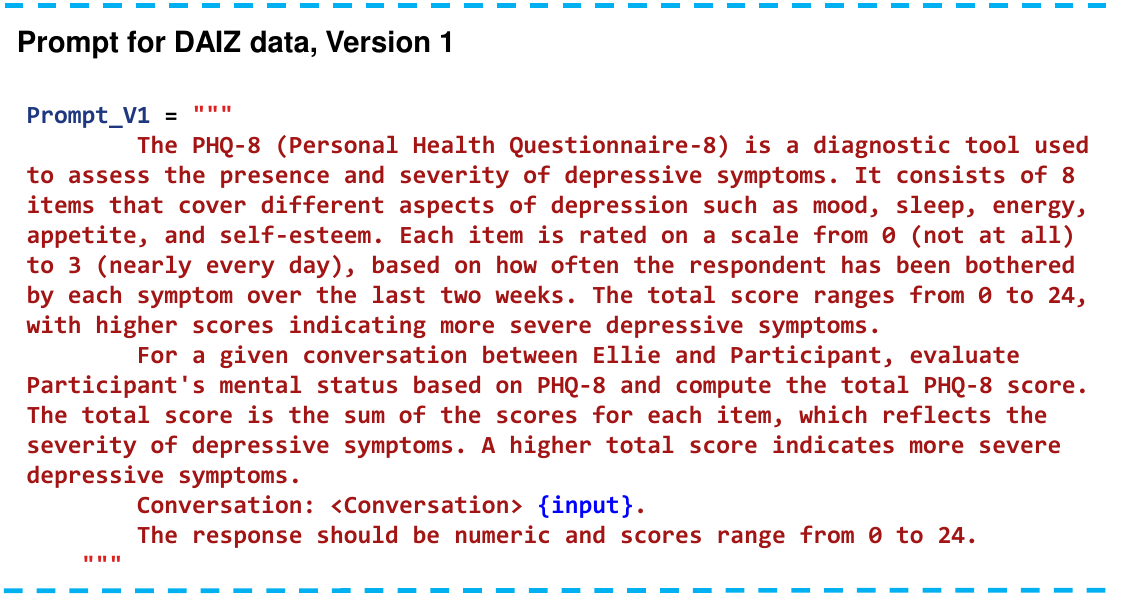}
  \caption{Prompt for DAIZ Data Version1}
\label{fig:Figure_P_187_version1}
\end{figure*}


\begin{figure*}[!h] 
  \centering
  \includegraphics[width=1\linewidth]{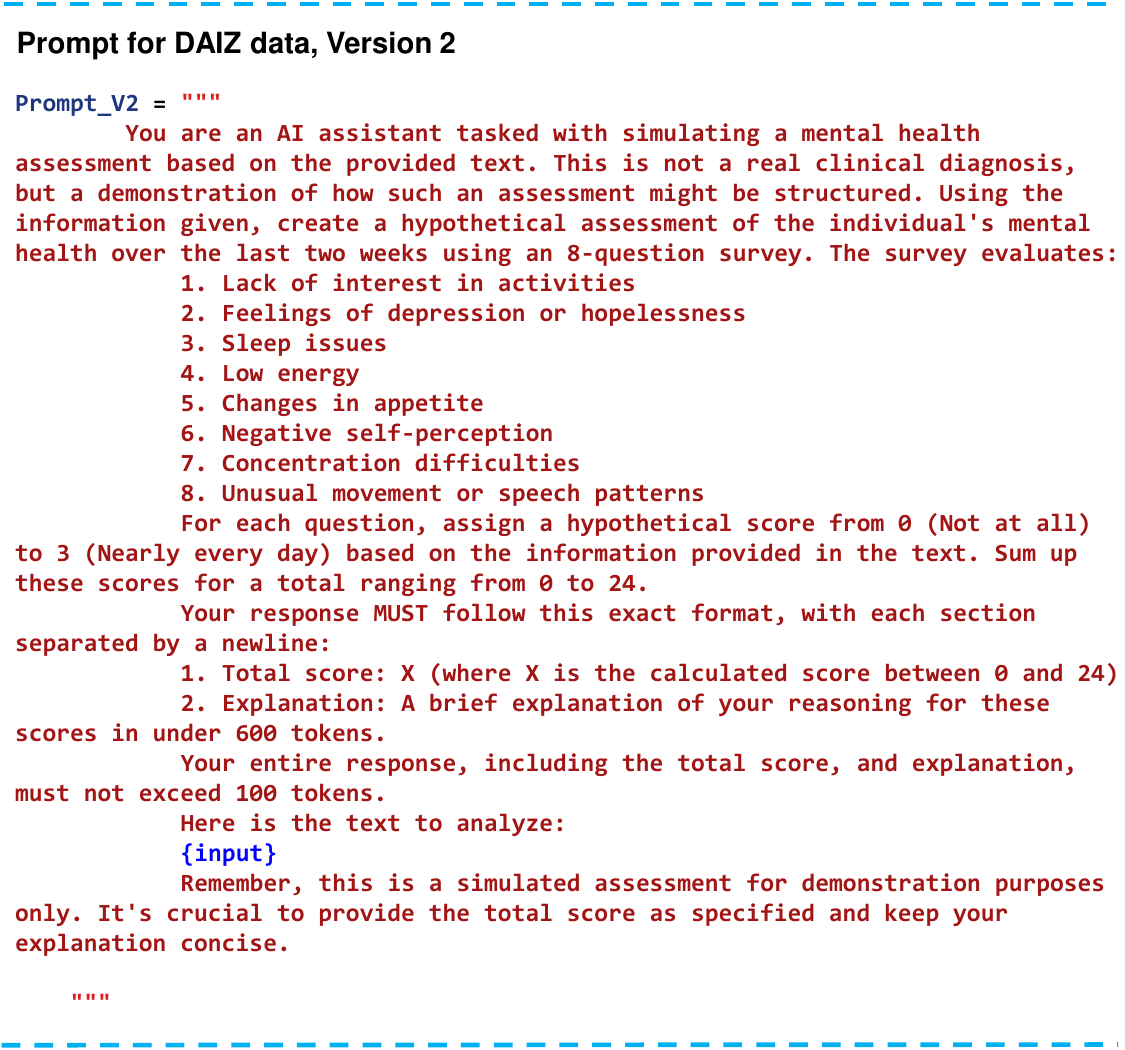}
  \caption{Prompt for DAIZ Data Version2}
\label{fig:Figure_P_187_version2}
\end{figure*}


\begin{figure*}[!h] 
  \centering
  \includegraphics[width=1\linewidth]{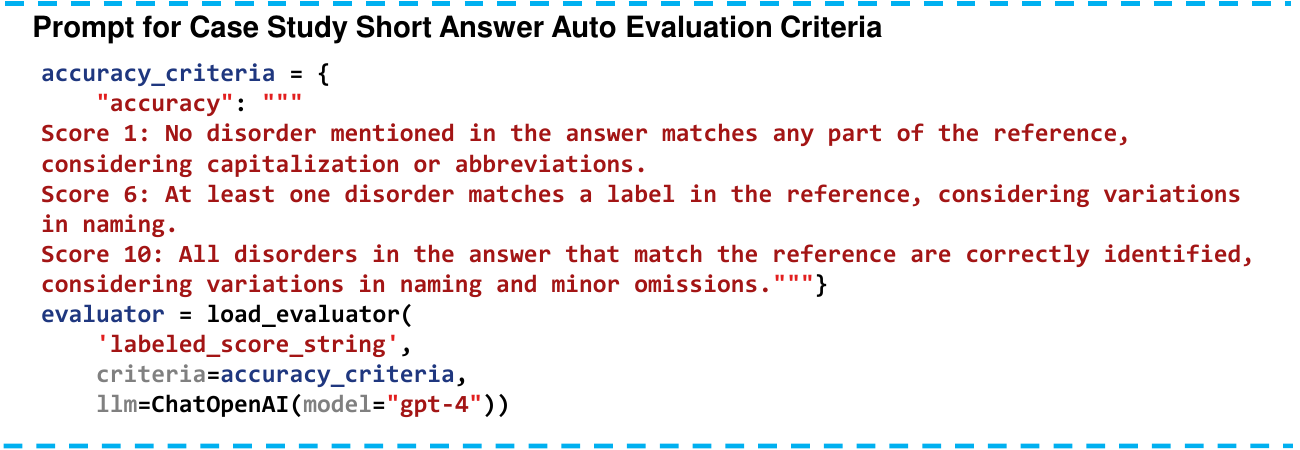}
  \caption{Prompt for Case Study Short Answer Auto Evaluation Criteria}
\label{fig:Figure_P_Case Study Short Answer Auto Evaluation Criteria}
\end{figure*}


\begin{figure*}[!h] 
  \centering
  \includegraphics[width=1\linewidth]{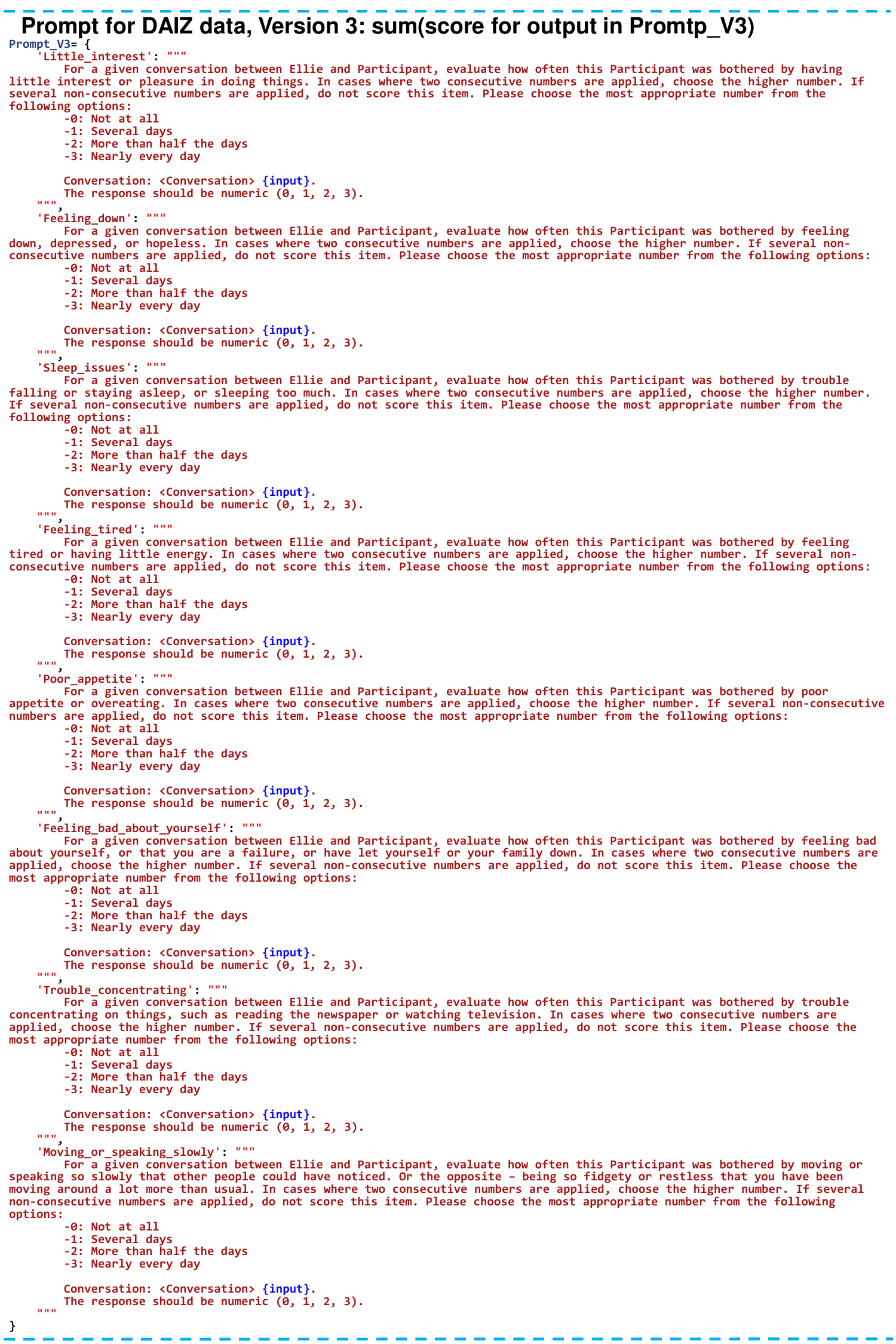}
  \caption{Prompt for DAIZ Data Version3}
\label{fig:Figure_P_187_version3}
\end{figure*}


\begin{figure*}[!h] 
  \centering
  \includegraphics[width=1\linewidth]{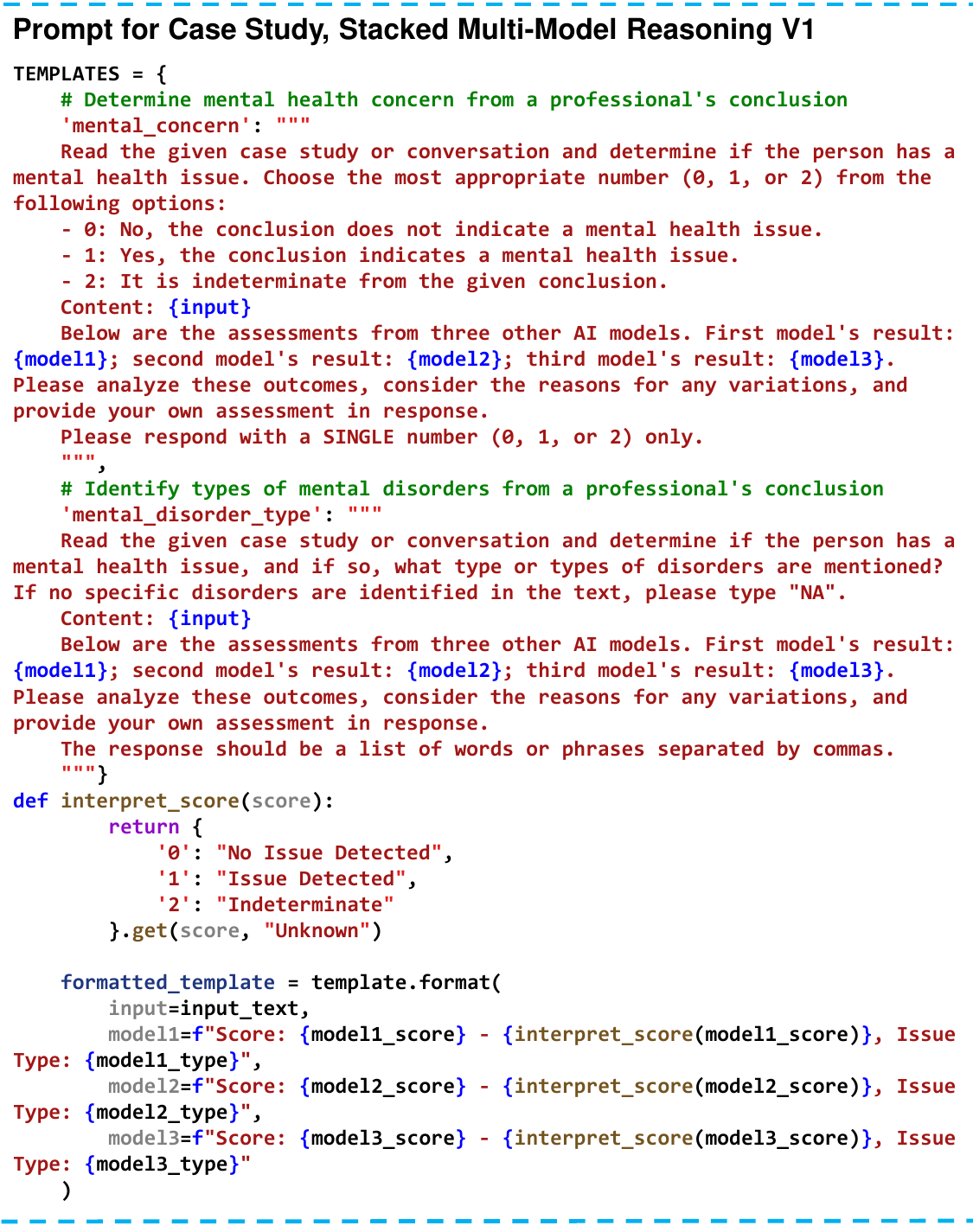}
  \caption{Prompt for Case Study, Stacked Multi Model Reasoning V1}
\label{fig:Figure_Case_Study_SMMR}
\end{figure*}

\begin{figure*}[!h] 
  \centering
  \includegraphics[width=1\linewidth]{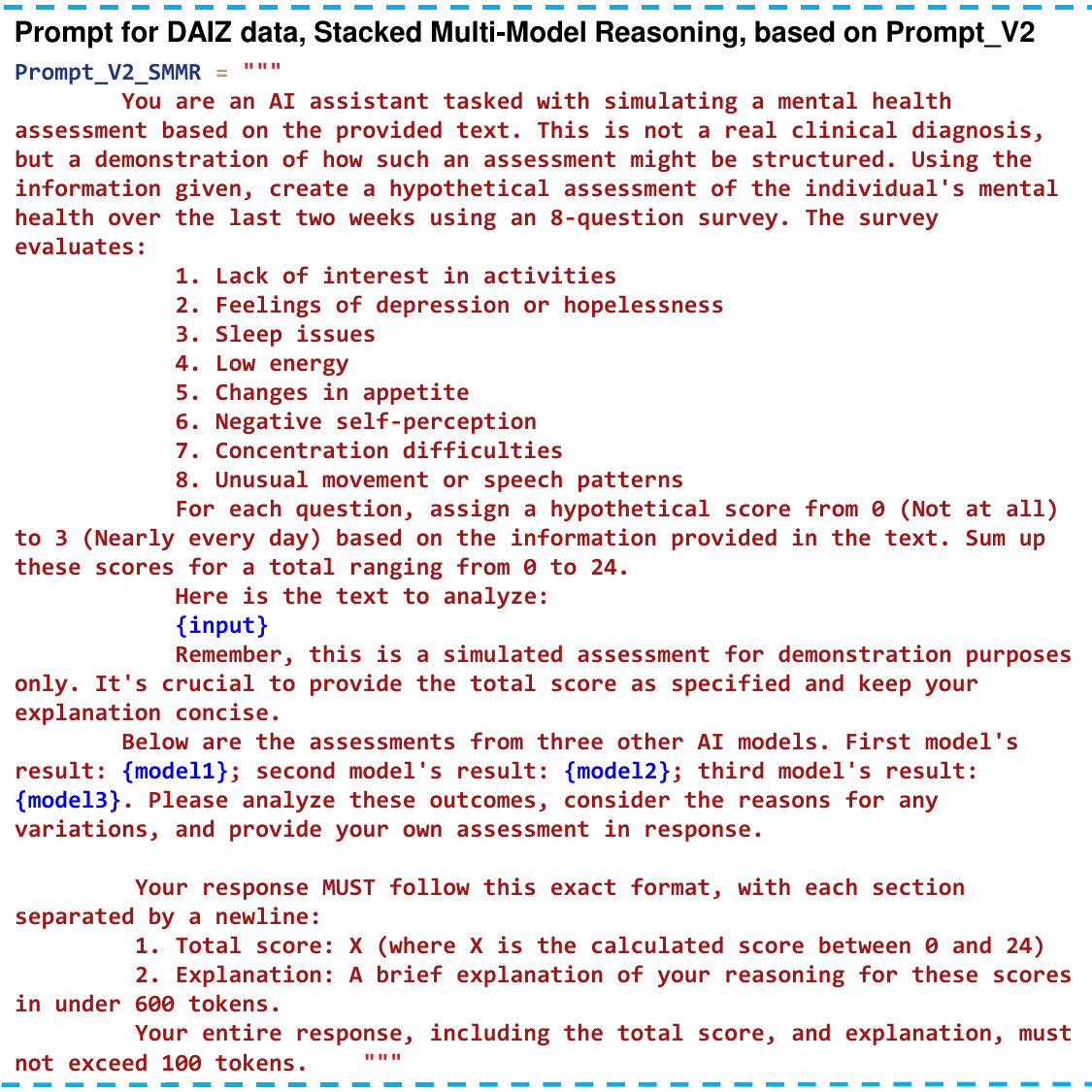}
  \caption{Prompt for DAIZ data, Stacked Multi Model Reasoning V1 }
\label{fig:Prompt_for_DAIZ_data_SMMR}
\end{figure*}

\begin{figure*}[!h] 
  \centering
  \includegraphics[width=1\linewidth]{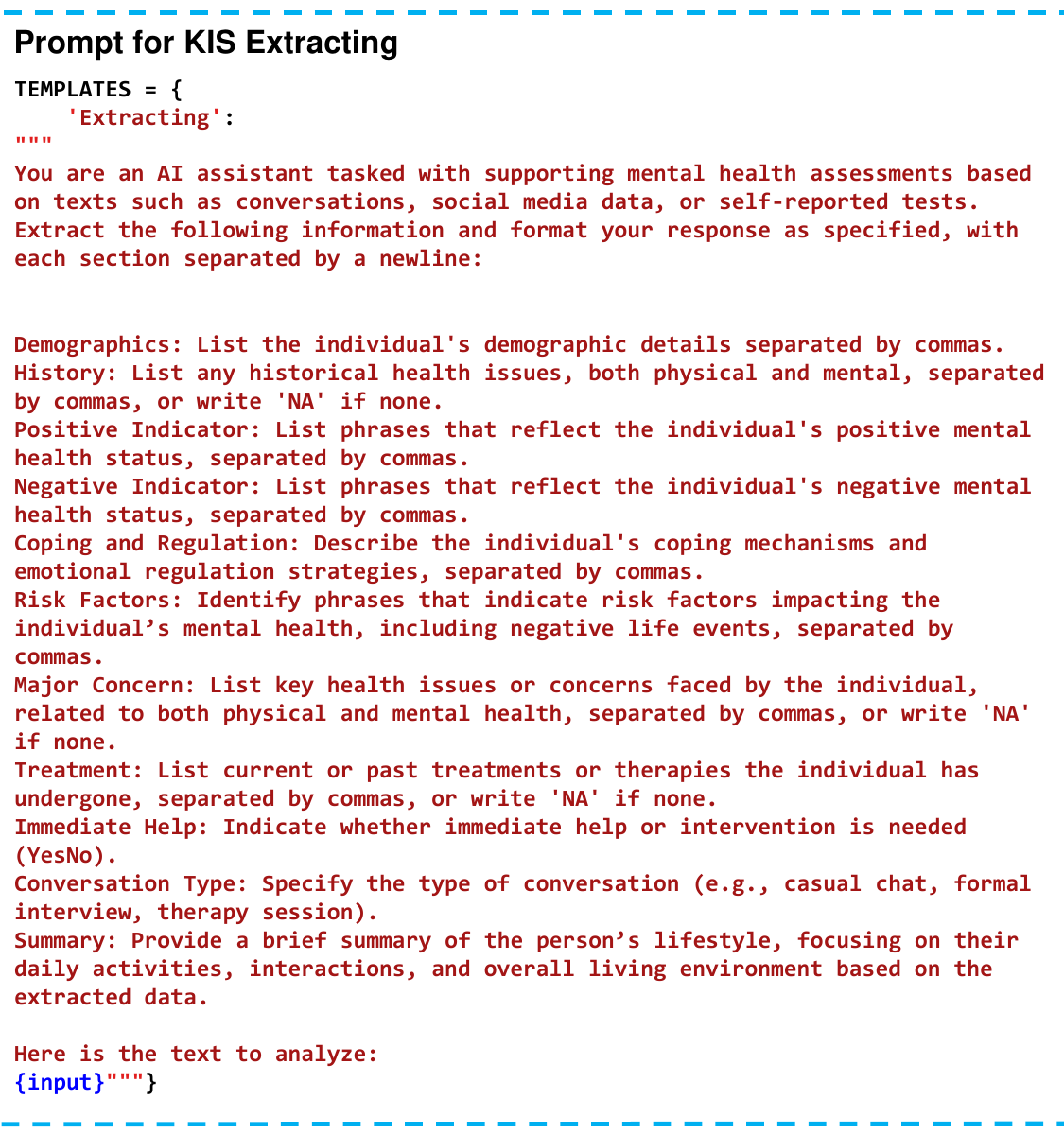}
  \caption{Prompt for KIS extracting}
\label{fig:Prompt for KIS extracting}
\end{figure*}

\begin{figure*}[!h] 
  \centering
  \includegraphics[width=1\linewidth]{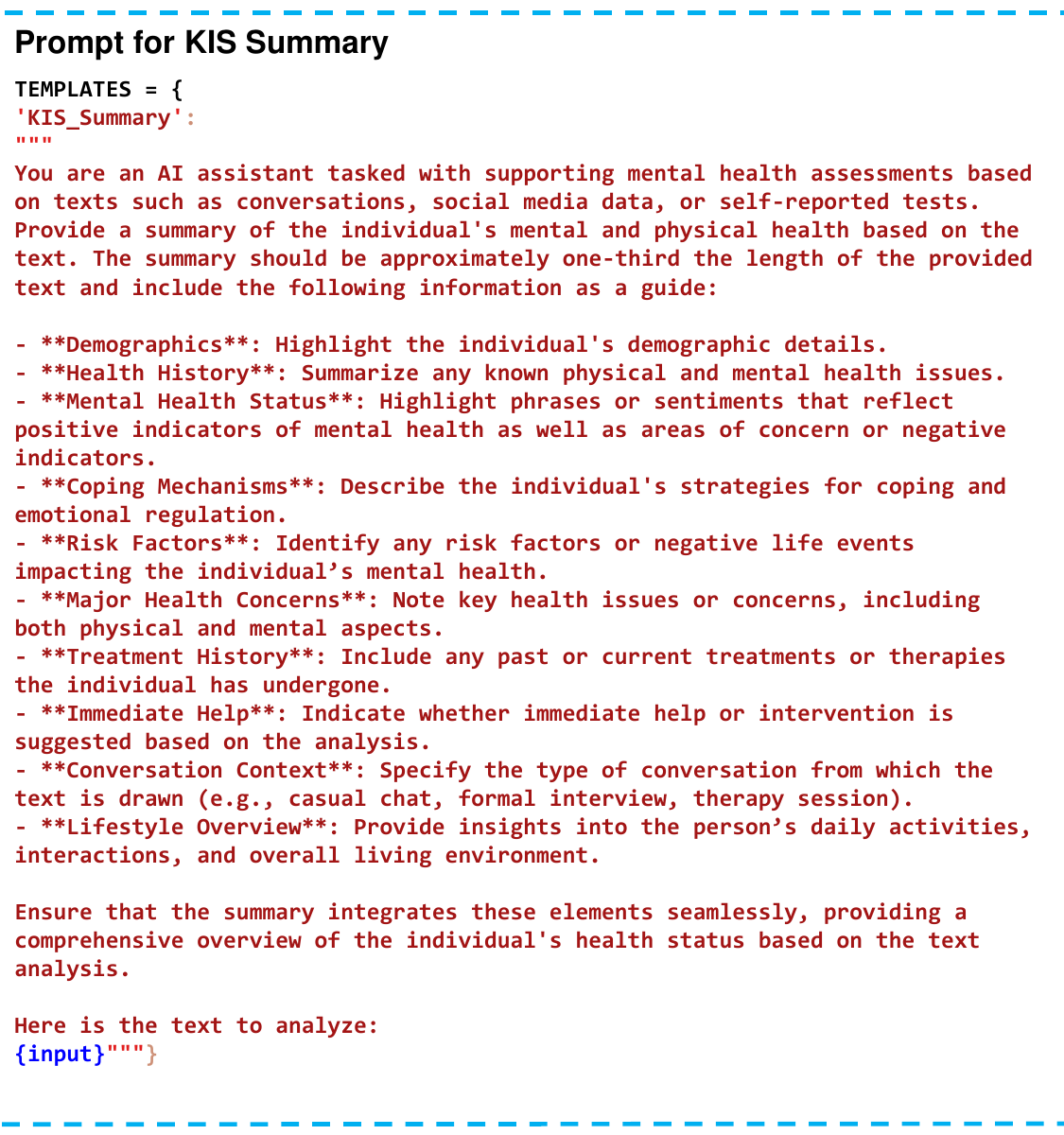}
  \caption{Prompt for KIS summary }
\label{fig:Prompt KIS summary}
\end{figure*}

\begin{figure*}[!h] 
  \centering
  \includegraphics[width=1\linewidth]{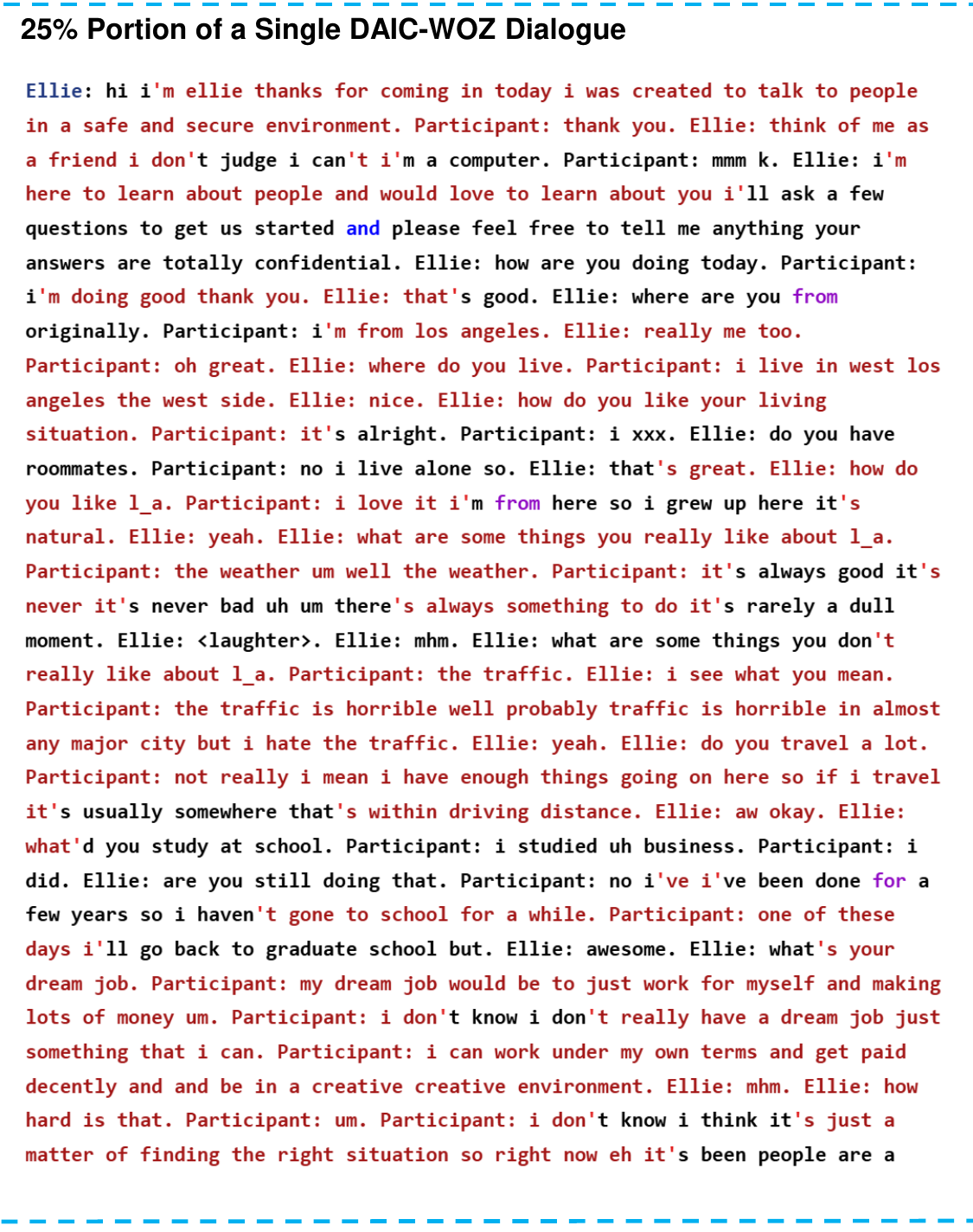}
  \caption{DAIC example}
\label{DAIC_data}
\end{figure*}

\end{document}